\documentclass{rtp_tech_report}

\usepackage{amsmath,amsfonts,bm}

\def\eqref#1{equation~\ref{#1}}

\def\1{\bm{1}}

\DeclareMathAlphabet{\mathsfit}{\encodingdefault}{\sfdefault}{m}{sl}
\SetMathAlphabet{\mathsfit}{bold}{\encodingdefault}{\sfdefault}{bx}{n}

\usepackage{url}

\title{Super-Resolution in the Right Latent Space: \\ A Frozen Vision-Foundation Substrate}

\author{%
\parbox{\dimexpr\textwidth-1pt\relax}{\centering
Wanzhou Lei$^{1,2,\sharp}$,
Cuifeng Shen$^{2}$,
Yanjin He$^{2,3,\sharp}$,
Maohua Li$^{2,4,\sharp}$,
Hua Yuan$^{5}$\\[1mm]
Per-Olof Persson$^{1}$,
Tao Lan$^{2}$,
Kan Liu$^{2}$,
Hanlin Tang$^{\dagger,2}$
}}

\affiliation{%
\parbox{\dimexpr\textwidth-1pt\relax}{\centering\small
$^1$University of California, Berkeley \quad $^2$Alibaba Group \quad $^3$University of Michigan, Ann Arbor \\ $^4$Nanjing University \quad $^5$Southeast University
}}
\renewcommand{\contributionlist}{\vspace{-4mm}}

\newcommand{\authorfootnotes}{%
  \begingroup
  \renewcommand{\thefootnote}{}%
  \footnotetext{\textsuperscript{$\dagger$}Project lead \quad $^\sharp$Work done during internship at Alibaba}%
  \endgroup
}
\checkdata[E-mail]{wanzhou\_lei@berkeley.edu, wanghaisheng.whs@alibaba-inc.com}

\abstract{
In an image latent space, the embeddings of high-resolution, natural, and sharp images form a manifold. Degradation of high-resolution images pushes their embeddings off this manifold. Real-world super-resolution (SR) then becomes the task of mapping the degraded embedding back onto this manifold: not anywhere on the manifold, but to the point that preserves what the input still carries, both its semantics and pixel details. Every published method implements this mapping in a reconstruction-oriented latent space or pixel space. We claim these spaces are the wrong substrates for SR. Low-resolution and degraded images are embedded far from the manifold, making the mapping difficult and expensive. The lack of semantic information in these substrates also makes it difficult to navigate to the faithful point on the manifold, causing severe hallucination when degradation is heavy. Thus, restoring in a suitable latent space is crucial
to the SR task. We show that the latent space of $23$ fused layers of a frozen DINOv3-L is one such space that makes the SR task easier. Degraded images are embedded near the manifold. Moreover, this substrate contains a hierarchy of information, from pixel record to degradation-robust semantics, directly guiding the model to find the faithful point on the manifold. On this substrate, a $415$M decoder is trained under reconstruction and adversarial objectives to map the degraded embeddings back and decode to pixel space in one pass. The resulting model, RAESR, attains the best fidelity--perception trade-off among state-of-the-art adversarial and diffusion-based restorers on RealSR, DRealSR, LSDIR and DIV2K-Val, at $37$\,ms per $512{\times}512$ image on a single H20 GPU. Swapping the substrate for a VAE latent under an identical recipe loses on
every metric.
}

\begin{document}
\maketitle
\authorfootnotes
\section{Introduction}
\label{sec:intro}
In an image latent space, the embeddings of clean, sharp, natural images form a manifold. Degradation pushes an image's embedding off this manifold, and real-world SR is the task of mapping it back onto the correct point on the manifold, the point that stays faithful to the input in both semantics and surviving detail. Choosing the suitable latent space as the substrate for this task is therefore crucial. A suitable space should offer two properties. First, degraded images should be embedded \emph{near} the manifold, so that the map back is easier and cheaper. Second, the space should carry a rich hierarchy of information to directly guide the map to the faithful point. High-level semantics tell the model which details are correct to synthesise, and low-level pixel information tells it which details must be preserved and what is noise to be removed.

\begin{figure}[t]
\centering
\includegraphics[width=\textwidth]{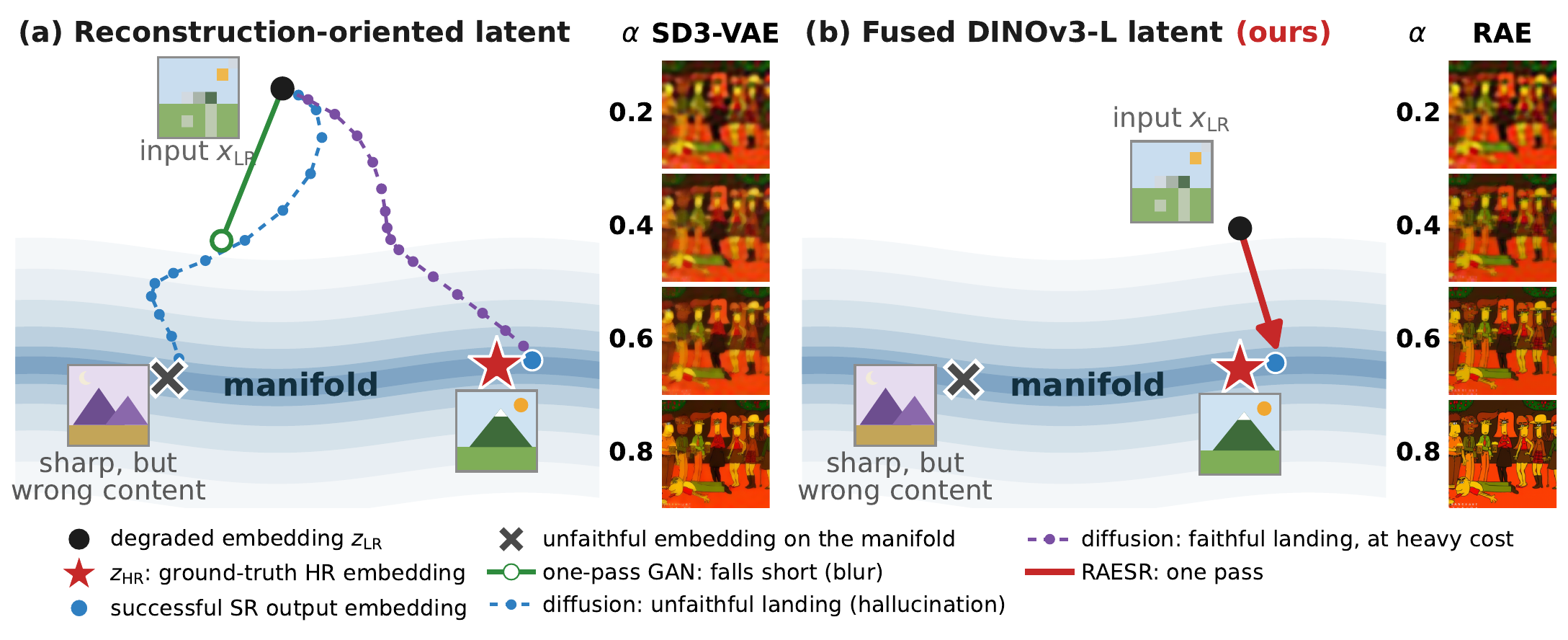}
\caption{\textbf{The choice of latent space decides the task.} \textbf{(a)} A
reconstruction-oriented latent. \textbf{(b)} The fused DINOv3-L latent (ours). \textbf{Strips:}
decoded images along the straight latent line
$z_\alpha = (1-\alpha)\,z_{\mathrm{LR}} + \alpha\,z_{\mathrm{HR}}$ in each space.}
\label{fig:intro-geo}
\end{figure}

Current real-world SR methods restore in spaces with neither property. They either restore in the pixel space itself \citep{realesrgan,bsrgan,hat}, or in the latent of a variational autoencoder \citep{stablesr,seesr,supir,osediff}. Since these spaces are reconstruction-oriented, they faithfully preserve the degradation. As a result, degraded images are embedded far from the manifold (Fig.~\ref{fig:intro-geo}a). The map to the manifold is therefore hard and expensive by default, and the lack of semantic information makes it even harder to navigate to the faithful point on the manifold. Lightweight GAN restorers in these spaces usually miss the manifold. Other methods make the distant map possible by training or adopting a visual generative prior such as diffusion, and pay with huge model size and slow inference. Even then, the long and weakly guided map can land on the manifold away from the faithful point, creating hallucinations (Fig.~\ref{fig:intro-geo}a). Mitigations have been tried: fine-tuning the encoder to close the distance between degraded and clean embeddings \citep{faithdiff}, and injecting extra semantic guidance, from a text encoder \citep{seesr,supir} or from the features of a vision foundation model \citep{vosr,f2idiff}. However, these never solve the problem itself. 

We believe SR must be conducted in a suitable latent space to be efficient, and we claim that the averaged layers of the DINOv3-L encoder form such a latent space (\S\ref{sec:contains}). DINOv3 \citep{dinov3} is a self-supervised vision transformer trained on $1.7$ billion curated, unlabelled images by self-distillation. Its network learns to match its own representations
across augmented views of the same image, rather than to reconstruct pixels. As a result, degraded images are encoded close to the manifold (Fig.~\ref{fig:intro-geo}b). The decoded images along the straight interpolation line between the degraded and the clean embedding show this directly
(Fig.~\ref{fig:intro-geo}, strips). In this latent space the decodes turn clear and natural much earlier than in the SD3-VAE latent (\S\ref{sec:geometry}). Moreover, the fusion preserves a hierarchy of information, from a near-lossless pixel record in the shallow layers to degradation-robust semantics in the deep layers, which provides direct guidance for the map to the faithful point (\S\ref{sec:contains}). Therefore, the one-step map is much easier to learn, as shown in Fig.~\ref{fig:intro-geo}b.

\begin{figure}[t]
\centering
\includegraphics[width=\textwidth]{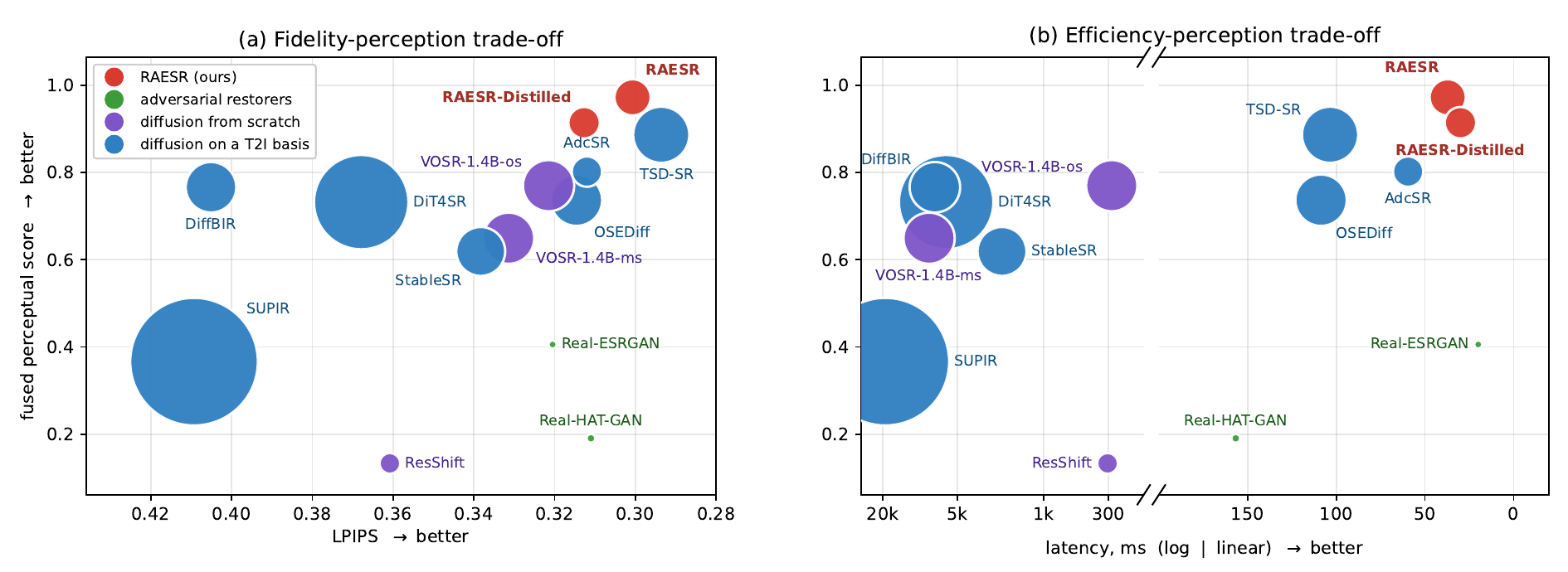}
\caption{\textbf{Perceptual-Fidelity and Perceptual-Efficiency tradeoffs.} Circle area scales with
inference parameter count. The vertical axis fuses the six perceptual metrics, each min--max normalized and re-oriented, then averaged with equal weight. \emph{(a)} against LPIPS; \emph{(b)} against measured latency (one H20, batch $1$). Construction and
robustness checks of the fused score: App.~\ref{app:fused}.}
\label{fig:tradeoff}
\end{figure}

\textbf{RAESR} is built on this substrate. One pass of this single trained $415$M decoder maps the degraded embedding back and decodes it to pixels. Averaged over four benchmarks, RAESR is first among the twelve published methods we evaluate on NIQE, MANIQA and DISTS and second on MUSIQ, LPIPS and TOPIQ (Table~\ref{tab:main}). It attains the best fidelity--perception trade-off and the best efficiency--perception trade-off (Fig.~\ref{fig:tradeoff}), at $719$M inference parameters and $37$\,ms per image on one H20. 

\paragraph{Contributions.}
\emph{(i)} We build RAESR, an efficient one-pass restorer that attains the state-of-the-art fidelity--perception and efficiency--perception trade-offs among the twelve published methods we evaluate, at $37$\,ms per image, with a distilled variant at $30$\,ms (\S\ref{sec:experiments},
App.~\ref{app:distill}).
\emph{(ii)} We redirect future SR research toward choosing a suitable latent substrate for the task.
\emph{(iii)} We characterize what makes a substrate suitable. Degraded images should be embedded near the manifold. All levels of information should be present in the substrate as direct guidance to the faithful point (\S\ref{sec:intro}).
\emph{(iv)} We rigorously analyze why the fused latent of DINOv3-L is such a space (\S\ref{sec:contains}, \S\ref{sec:analysis}).

\section{Related Work}
\label{sec:related}
\paragraph{Restoring in a reconstruction-oriented space.}
Every line of real-world SR operates in such a space. Since these spaces offer neither of the two aforementioned properties, each family pays its own price.  The one-pass adversarial restorers Real-ESRGAN \citep{realesrgan}, BSRGAN \citep{bsrgan} and Real-HAT-GAN \citep{hat} perform the map directly in one step, with no generative prior or external guidance. They are small and fast, but they fall short of the manifold and outputs are blurry. The diffusion line makes the long map feasible with a pretrained text-to-image prior. StableSR
\citep{stablesr}, DiffBIR \citep{diffbir}, SeeSR \citep{seesr}, PASD \citep{pasd}, SUPIR
\citep{supir} and DiT4SR \citep{dit4sr} sample it over tens of steps, at billions of parameters
and seconds per image (Table~\ref{tab:main}). OSEDiff \citep{osediff}, PiSA-SR \citep{pisasr},
TSD-SR \citep{tsdsr} and AdcSR \citep{adcsr} distil the schedule to one step but keep the
backbone, while ResShift \citep{resshift} and SinSR \citep{sinsr} train a short chain from
scratch. None of them changes substrate, so the map stays long and weakly guided.

\paragraph{Working around the two defects.}
There have also been attempts to mitigate the two defects of the reconstruction-oriented latent space. Against the distance,
FaithDiff \citep{faithdiff} fine-tunes the VAE encoder jointly with the prior to close the gap
between degraded and clean latents. DiffBIR \citep{diffbir} restores in pixel space before
generating. Against the missing direct guidance, StableSR \citep{stablesr} trains a time-aware encoder
to condition the frozen prior, SeeSR \citep{seesr} conditions on tags from a degradation-aware
extractor, PASD \citep{pasd} adds a captioner and pixel-aware cross-attention, SUPIR
\citep{supir} scales to large-scale text annotation, and, nearest to this work, VOSR
\citep{vosr} and F2IDiff \citep{f2idiff} condition the generator on DINOv2 features. Every remedy still preserves the reconstruction-oriented substrate. We take the opposite route and change the substrate.

\paragraph{Representation autoencoders.}
RAE \citep{rae} and RAEv2 \citep{raev2} show that a frozen self-supervised encoder with a
trained decoder can replace the VAE for diffusion transformers, and RAEv2 introduces the
multi-layer fusion our latent adopts (\S\ref{sec:contains}). That line is developed for
generation and clean reconstruction. We carry the
substrate to restoration.

\section{Method}
\label{sec:method}

\subsection{Fused Layers of DINOv3-L as the Latent}
\label{sec:contains}
\label{sec:fusion}

\begin{figure}[t]
\centering
\includegraphics[width=\textwidth]{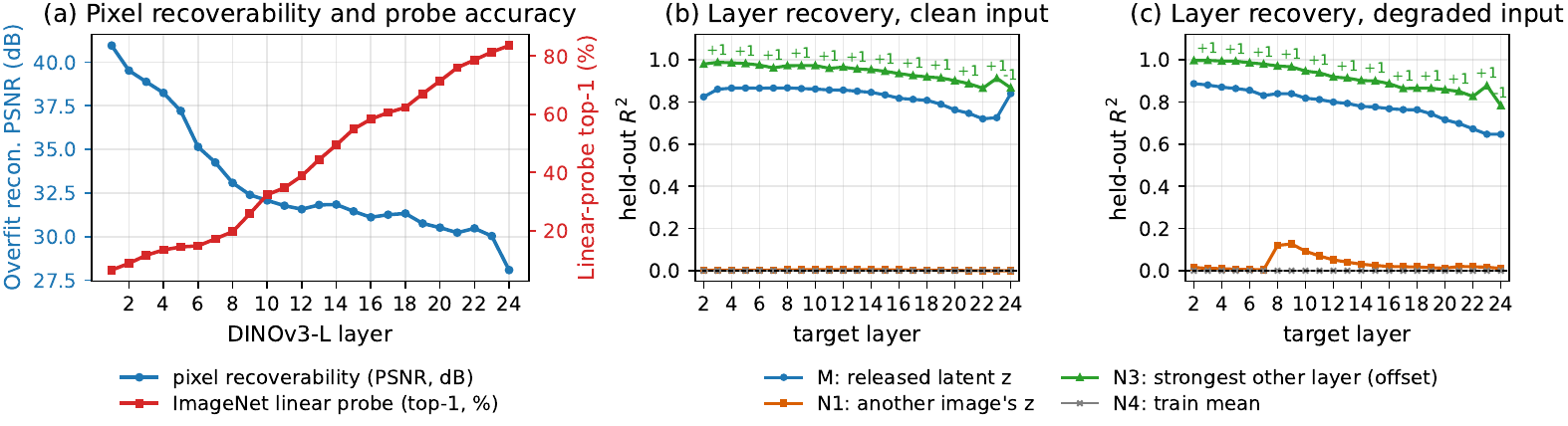}
\caption{\textbf{(a)} Per layer, clean inputs: PSNR of a small decoder overfit to a few images
(blue) and held-out ImageNet linear-probe accuracy (red). \textbf{(b, c)} $R^2$ of a linear map
from the fused latent $z$ to each layer, on clean and training-degraded inputs.}
\label{fig:geom-depth}
\end{figure}

The $24$ layers of DINOv3-L provide a hierarchy of information, from pixel details to abstract semantics. We take each layer and overfit a small decoder on its tokens to reconstruct the input images. The reconstruction PSNR decreases from $40.95$\,dB at the first layer to $28.09$\,dB at the last. On the same layers, the top-1 accuracy of a linear ImageNet probe rises from $6.5\%$ to $83.6\%$ (Fig.~\ref{fig:geom-depth}a). This hierarchy of information provides useful and direct guidance for the map to the faithful point on the manifold.  RAEv2 forms its latent from the average of the last $K$ layers plus the final layer's spatial average. We follow that construction with $K = 23$, so that only the first layer is left out:
\begin{equation}
  z \;=\; \frac{1}{23}\sum_{\ell=2}^{24} h_\ell
      \;+\; \mathrm{broadcast}\Big(\tfrac{1}{N}\textstyle\sum_{p} h_{24}(p)\Big),
  \qquad h_\ell \in \mathbb{R}^{N\times d}.
  \label{eq:agg}
\end{equation}

Averaging destroys little information in each layer. Analysis shows that distant layers
have disentangled subspaces and neighbouring layers share redundancy; both factors minimise the loss of information when taking the mean (App.~\ref{app:disentangle}). We also fit a linear map from the fused latent $z$ to recover each layer on clean and training-degraded inputs. Note that the linear map's recovery ability is a lower bound on the highly non-linear decoder's recovery ability since it can only extract information that is linearly available. Yet, the linear map still achieves high $R^2$. Most layers at about $0.8$ or higher and the shallow layers near $0.9$ (Fig.~\ref{fig:geom-depth}b,c). More
rigorous analysis is in App.~\ref{app:recovery}.

\subsection{One decoder pass}
\label{sec:pipeline}
Because the degraded images are encoded near the manifold and the latent carries a rich hierarchy of information, the mapping back is short and well-guided. Thus, a single medium-sized decoder suffices to make it and to decode to the pixel space in one pass. Let $x_{\mathrm{LR}}$ be a degraded and low-resolution observation of a high-resolution image $x_{\mathrm{HR}}$. We place the entire restoration burden on a trainable decoder $G_\theta$ over the frozen substrate:
\begin{equation}
  \hat{x}_{\mathrm{HR}} \;=\; G_\theta\!\left(E(Upsample(x_{\mathrm{LR}}))\right),
  \label{eq:pipeline}
\end{equation}
where \emph{Upsample} denotes the bicubic upsampling operation to the same dimension as
$x_{\mathrm{HR}}$. There is no sampling loop and no iterative refinement. After the decoder we align
each output channel's mean to the input's, which avoids any colour shift and is a free improvement
at almost no cost (App.~\ref{app:perset}). The decoder $G_\theta$ embeds the latent with a linear
map $d \to 1152$, applies $28$ transformer layers ($16$ heads, MLP width $4096$, 2D RoPE
\citep{rope}) over the token grid, and predicts $16^2{\times}3$ pixels per token before
unpatchifying, with a total of $415$M parameters.

\subsection{Training objectives}
\label{sec:objective}

The decoder must map onto the manifold, at the point faithful to the input. It must
reproduce what survived the degradation and synthesise what the degradation destroyed in the right details. The loss
emphasizes both tasks. The encoder is frozen throughout, in each stage, only $G_\theta$ was trained to minimize
\begin{equation}
  \mathcal{L} \;=\; \mathcal{L}_{1}
  \;+\; \lambda_{\mathrm{p}}\,\mathcal{L}_{\mathrm{LPIPS}}
  \;+\; \lambda_{\mathrm{adv}}\,\mathcal{L}_{\mathrm{adv}},
  \label{eq:loss}
\end{equation}
with $\lambda_{\mathrm{p}}\!=\!1.0$ (VGG16 LPIPS \citep{lpips}) and $\lambda_{\mathrm{adv}}\!=\!0.5$ wherever the adversarial term is active. Both $\mathcal{L}_1$ and LPIPS only supervise faithfulness and cannot map the embedding back onto the manifold. Where details are destroyed, the reference is one completion among many valid ones, and a loss that
regresses to the reference optimizes to the average of those completions, which is blur \citep{blau2018}. The adversarial term supplies what regression cannot. It accepts any completion whose texture is real rather than pulling toward the mean, and so pushes the embedding onto the manifold. It is also the part of the loss that needs the most care
(\S\ref{sec:critic}).

Training proceeds in three stages. Stage~1 starts from the official RAEv2 ViT-XL reconstruction
decoder \citep{raev2}, replaces its original positional embedding with 2D RoPE, and trains
without the adversarial term on bicubically downsampled $512$px LSDIR images. Stage~2 applies strong second-order Real-ESRGAN degradation \citep{realesrgan} at $512$px,
and trains with the critic active. Stage~3 aligns training with deployment. It is a short finish under a mixture
of strong, medium and mild degradations. App.~\ref{app:training} documents data, schedules and every hyper-parameter.

\subsection{Adversarial training}
\label{sec:critic}

\begin{figure}[t]
\centering
\includegraphics[clip, trim=8 16 46 54, width=\textwidth]{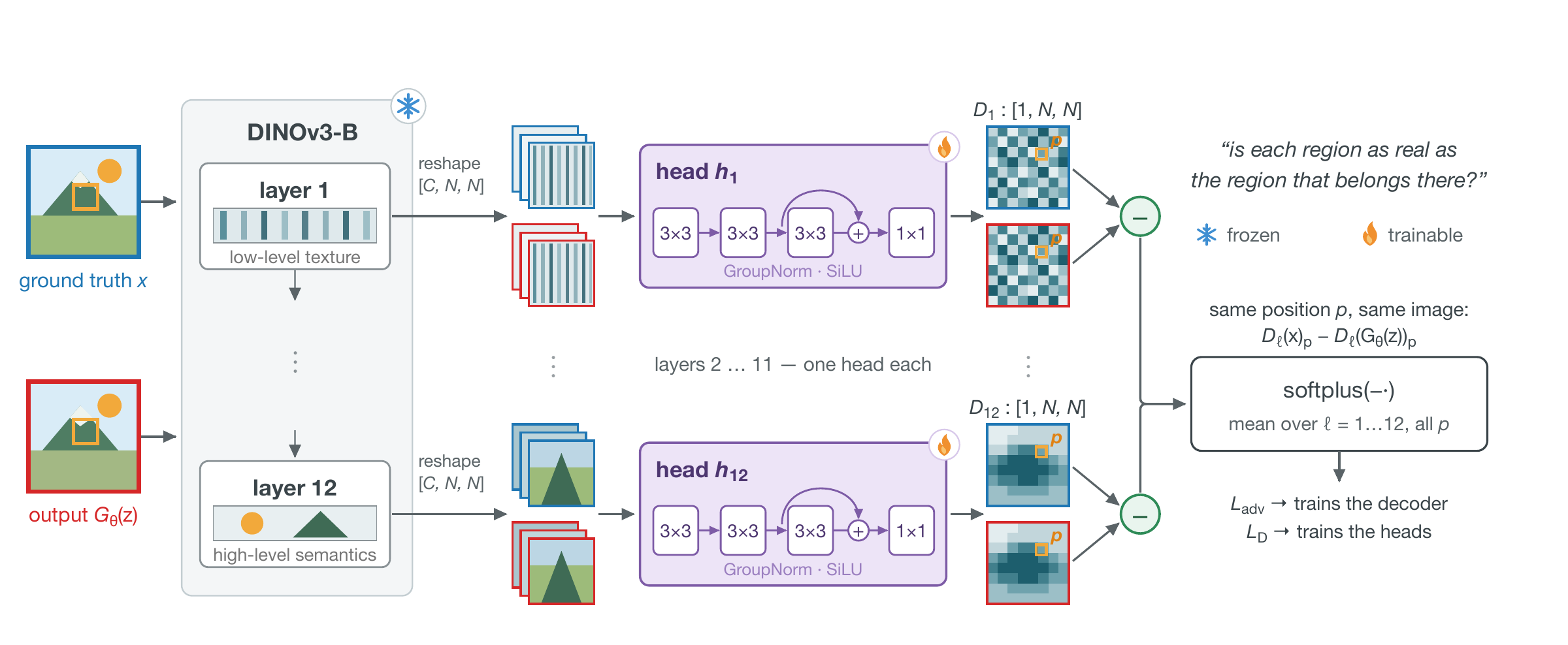}
\caption{\textbf{The critic.} The restored image and its ground truth pass through the same
frozen DINOv3-B, and each of its $12$ layers is read by its own trainable head that emits one
logit per token.}
\label{fig:critic}
\end{figure}

The adversarial term is what guides the learning of the map back onto the manifold. Thus, we need a critic that can provide rich and direct guidance. It needs the same hierarchy of information to decide whether a synthesised region is right for the image at each level, from low-level texture statistics to high-level semantics. Following the projected discriminators of \citet{projdisc} and the off-the-shelf critics of \citet{visionaided}, we therefore build it on the same representation family the substrate is built from (Fig.~\ref{fig:critic}).  The restored image and its ground truth pass through a frozen DINOv3-B encoder, and each of its $12$ layers is read by its own trainable convolutional head, which emits one logit per token as the ``realness'' of that token in that particular information level. The heads carry $191$M learnable parameters in total and are discarded after training. Reading every layer through its own head is the critic-side counterpart of fusing every layer into $z$. Realism is judged at every level, from texture statistics to the plausibility of semantics.

Two properties of this design, both visible in Fig.~\ref{fig:critic}, make that realism
requirement fine-grained and relative. First, the judgment is \emph{dense}. No logit is pooled into an image-level score, so
every region is judged on its own and a weak region cannot be averaged away by the rest. Second, it is
\emph{paired by position}. The logit of the restoration at position $p$ is compared with the logit
of the ground truth at the same position of the same image, never with a batch statistic. The bar at each position is therefore the realism the reference itself attains there. Where the reference is smooth the bar is low, so the term does not ask for texture the scene does not have. Where the
reference is detailed the bar is high, and the requirement is stricter. 

The training objective is that of Adversarial Flow Models \citep{afm}. Write $r$ and $f$ for the
real and generated images, $D_\ell(\cdot)_p$ for the logit of head $\ell$ at position $p$,
$\mathrm{sg}$ for the stop-gradient, and $\mathbb{E}_{\ell,p}$ for the mean over the $12$ heads
and all token positions. Training alternates between the two parameter sets. The critic heads
minimize
\begin{align}
  \mathcal{L}_{D} &= \mathbb{E}_{\ell,p}\big[\mathrm{softplus}\big(-(D_\ell(r)_p - D_\ell(f)_p)\big)\big]
     \;+\; \lambda_{\mathrm{gp}}\,(R_1 + R_2) \;+\; \lambda_{\mathrm{cp}}\,\mathrm{CP},
     \label{eq:dloss}
\end{align}
and the decoder minimizes Eq.~\ref{eq:loss}, whose adversarial term, computed with the frozen critic, is
\begin{align}
  \mathcal{L}_{\mathrm{adv}} &= \mathbb{E}_{\ell,p}\big[\mathrm{softplus}\big(-(D_\ell(f)_p - \mathrm{sg}[D_\ell(r)_p])\big)\big].
  \label{eq:gloss}
\end{align}
The loss is relativistic \citep{rgan}. The critic is trained to score the ground truth above the
restoration, never to classify either in isolation. Three penalties, acting on the critic only,
follow \citet{afm}, at the weights of their released configuration
($\lambda_{\mathrm{gp}}\!=\!0.25$, $\lambda_{\mathrm{cp}}\!=\!0.01$).
$R_1 = \mathbb{E}_{\ell,p}\|\nabla_r D_\ell(r)_p\|^2$ and
$R_2 = \mathbb{E}_{\ell,p}\|\nabla_f D_\ell(f)_p\|^2$ are the zero-centred gradient penalties of
\citet{r1penalty}, carried on both sides as in \citet{r3gan} and evaluated by the
finite-difference approximation of \citet{apt}, so no second backward pass through the critic is
needed. The centring penalty
$\mathrm{CP} = \mathbb{E}_{\ell,p}(D_\ell(r)_p + D_\ell(f)_p)^2$ closes the one degree of freedom
the other terms leave open. Every one of them depends only on differences of logits, so the pair
could drift to arbitrarily large values together, and $\mathrm{CP}$ pins their sum near zero without touching the difference the losses train on \citep{afm}. App.~\ref{app:adv} documents the critic's training schedule, and App.~\ref{app:training} gives every hyper-parameter. 

\section{Experiments}
\label{sec:experiments}
\subsection{Protocol}
\label{sec:protocol}
We evaluate $4\times$ real-world SR on four benchmarks: RealSR \citep{realsr} ($100$ images), DRealSR \citep{drealsr} ($93$), LSDIR \citep{lsdir} ($250$) and DIV2K-Val \citep{div2k} (3000). None of them is seen in training. We compare twelve public methods (Table~\ref{tab:main}) that fall into three categories: adversarial restorers, diffusion models trained from scratch, and diffusion models built on a text-to-image basis. Every row is computed with the official checkpoint at official defaults. No method trains against any reported metric (App.~\ref{app:perset}). The cost columns of
Table~\ref{tab:main} are measured on one NVIDIA H20 at batch $1$ on the $128{\times}128
\rightarrow 512{\times}512$ task. Latency is the mean of $30$ CUDA-event measurements after $8$
warm-up iterations, memory is the peak GPU allocation reported by PyTorch over the timed run, and
parameters count every module on the timed path. Every method runs its official repository
unmodified at native precision. We report nine metrics,
PSNR, SSIM, LPIPS, DISTS, NIQE, MUSIQ, MANIQA, TOPIQ and CLIP-IQA
\citep{lpips,dists,niqe,musiq,maniqa,topiq,clipiqa}, averaged over the four benchmarks.

\subsection{Main results}
\label{sec:main-results}
\begin{table}[p]
\centering
\setlength{\abovecaptionskip}{3pt}\setlength{\belowcaptionskip}{3pt}
\renewcommand{\arraystretch}{0.92}
\footnotesize
\setlength{\tabcolsep}{1.3pt}
\caption{\textbf{$\times4$ real-world SR, macro-averaged over the four benchmarks}. NFE: network evaluations per image. RAESR$^{\rm D}$: distilled. \textbf{Bold} best, \underline{underline} second; per-benchmark tables in App.~\ref{app:perset}.}
\label{tab:main}
\resizebox{\textwidth}{!}{%
\begin{tabular}{lcccccccccc|rrr}
\toprule
Method & NFE & PSNR$\uparrow$ & SSIM$\uparrow$ & LPIPS$\downarrow$ & DISTS$\downarrow$ & NIQE$\downarrow$ & MUSIQ$\uparrow$ & MANIQA$\uparrow$ & TOPIQ$\uparrow$ & CLIP$\uparrow$ & \shortstack{Params\\(M)} & \shortstack{Lat.\\(ms)} & \shortstack{Mem\\(GiB)} \\
\midrule
\textbf{RAESR} & 1 & 24.08 & 0.6269 & \underline{0.3007} & \textbf{0.1970} & \textbf{4.319} & \underline{70.51} & \textbf{0.6713} & \underline{0.6815} & 0.7049 & 719 & 37.2 & 1.57 \\
\textbf{RAESR}$^{\rm D}$ & 1 & 23.39 & 0.6245 & 0.3127 & 0.2097 & \underline{4.485} & 70.44 & \underline{0.6652} & 0.6755 & 0.7033 & 502 & 30.0 & 1.18 \\
\midrule
\multicolumn{14}{l}{\emph{Adversarial restorers}} \\
Real-ESRGAN & 1 & \underline{24.46} & \underline{0.6648} & 0.3205 & 0.2208 & 5.282 & 60.50 & 0.5388 & 0.5270 & 0.5068 & 17 & 20.0 & 0.17 \\
Real-HAT-GAN & 1 & \textbf{25.56} & \textbf{0.6829} & 0.3110 & 0.2270 & 5.837 & 55.56 & 0.5310 & 0.4433 & 0.4087 & 21 & 156.7 & 0.73 \\
\midrule
\multicolumn{14}{l}{\emph{Diffusion trained from scratch}} \\
ResShift & 4 & 24.32 & 0.6396 & 0.3608 & 0.2440 & 7.471 & 56.69 & 0.5042 & 0.4768 & 0.5475 & 174 & 304.0 & 2.61 \\
VOSR-1.4B-ms & 25 & 23.87 & 0.6208 & 0.3314 & 0.2084 & 5.818 & 67.40 & 0.6240 & 0.6043 & 0.5540 & 1,743 & 8,454 & 7.23 \\
VOSR-1.4B-os & 1 & 23.76 & 0.6152 & 0.3215 & 0.2057 & 5.159 & 70.21 & 0.6399 & 0.6310 & 0.5805 & 1,746 & 281.4 & 7.24 \\
\midrule
\multicolumn{14}{l}{\emph{Diffusion on a T2I basis}} \\
AdcSR & 1 & 24.14 & 0.6309 & 0.3120 & 0.2087 & 4.918 & 68.60 & 0.6112 & 0.6651 & 0.6809 & 456 & 59.5 & 2.24 \\
StableSR & 20 & 23.74 & 0.6073 & 0.3383 & 0.2213 & 5.286 & 64.32 & 0.6026 & 0.5834 & 0.6476 & 1,555 & 2,175 & 7.26 \\
DiffBIR & 50 & 23.14 & 0.5534 & 0.4051 & 0.2412 & 5.356 & 69.53 & 0.6448 & \textbf{0.6961} & \underline{0.7193} & 1,683 & 7,593 & 4.97 \\
OSEDiff & 1 & 24.07 & 0.6417 & 0.3146 & 0.2127 & 5.186 & 67.91 & 0.6135 & 0.6206 & 0.6714 & 1,761 & 108.6 & 4.02 \\
TSD-SR & 1 & 23.51 & 0.6141 & \textbf{0.2936} & \underline{0.2026} & 4.733 & \textbf{70.95} & 0.6190 & 0.6613 & \textbf{0.7309} & 2,207 & 103.5 & 4.95 \\
DiT4SR & 40 & 22.08 & 0.5636 & 0.3679 & 0.2256 & 5.584 & 68.61 & 0.6573 & 0.6298 & 0.6667 & 8,381 & 6,152 & 17.12 \\
SUPIR & 50 & 23.40 & 0.5623 & 0.4093 & 0.2531 & 7.166 & 60.36 & 0.5770 & 0.5792 & 0.6716 & 18,152 & 19,112 & 46.79 \\
\bottomrule
\end{tabular}}
\end{table}

\begin{figure}[p]
\centering
\setlength{\abovecaptionskip}{3pt}
\includegraphics[width=\textwidth]{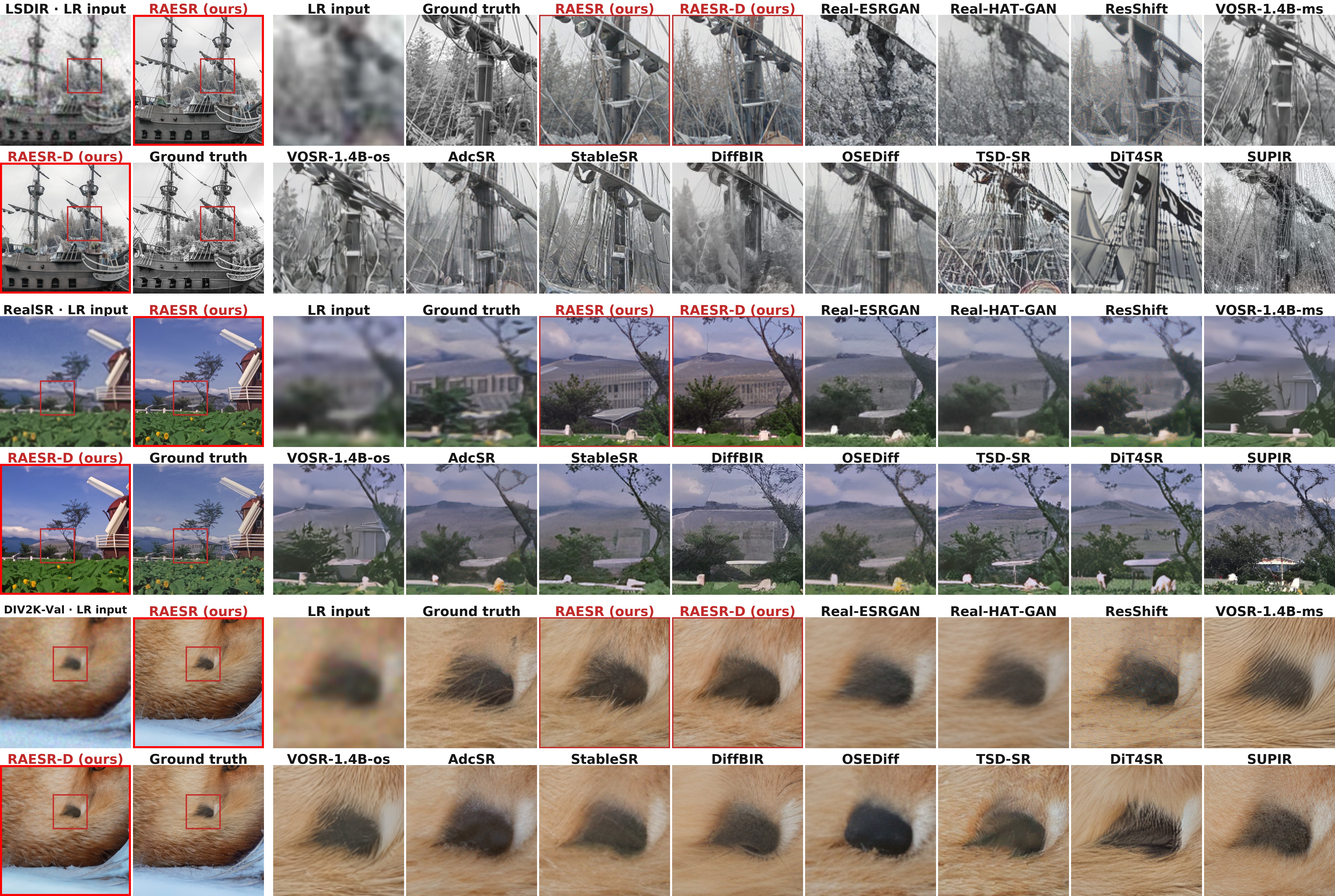}
\caption{\textbf{Qualitative $\times4$ comparison, all sixteen panels.} Full frames with the region boxed; right: that region
across every method at $4\times$ nearest-neighbour magnification, ours in \textcolor{red}{red}.
Blocks from top: LSDIR, RealSR, DIV2K-Val. \textbf{Best viewed zoomed in.} App.~\ref{app:gallery} shows other examples.}
\label{fig:qualitative}
\end{figure}

\begin{table}[!t]
\centering
\setlength{\abovecaptionskip}{3pt}\setlength{\belowcaptionskip}{3pt}
\caption{\textbf{Human preference, two-alternative forced choice with input and ground
truth shown.}}
\label{tab:human}
\small
\begin{tabular}{lcccc}
\toprule
Comparison & RAESR votes & Baseline votes & Ties & RAESR preference \\
\midrule
RAESR vs.\ Real-HAT-GAN & $163$ & $26$ & $26$ & $86.2\%$ \\
RAESR vs.\ DiffBIR      & $133$ & $55$ & $30$ & $70.7\%$ \\
RAESR vs.\ SUPIR        & $109$ & $80$ & $22$ & $57.7\%$ \\
RAESR vs.\ TSD-SR       & $101$ & $91$ & $27$ & $52.6\%$ \\
\bottomrule
\end{tabular}
\end{table}

The four-benchmark average is in Table~\ref{tab:main} and the per-benchmark tables in
App.~\ref{app:perset}. \textbf{RAESR achieves
the best scores among all twelve published methods on NIQE, DISTS and MANIQA}, and the second-best
on LPIPS, MUSIQ and TOPIQ.

Following the perception--distortion framework of \citet{blau2018}, we divide the nine metrics
into two families. PSNR, SSIM and LPIPS are fidelity metrics and measure the discrepancy between
the output and its reference. DISTS, NIQE, MUSIQ, MANIQA, TOPIQ and CLIP-IQA are perceptual
metrics and rate the perceived quality of the output as an image in its own right. Within the
fidelity family we place the most weight on LPIPS, which is calibrated on human judgments of
which of two images is closer to the reference \citep{lpips}. PSNR and SSIM instead reward exact
pixel agreement, and the degradation has made exact agreement impossible. The estimator that maximises PSNR is the blurred posterior mean rather than any sharp image \citep{blau2018,srgan}. Under this classification RAESR
attains the best fidelity--perception trade-off in the table. It is second on LPIPS, narrowly
behind TSD-SR, first on three of the six perceptual metrics with a clear margin on NIQE, and
second on two more. RAESR also attains the best trade-off between inference cost and performance
(Fig.~\ref{fig:tradeoff}). It runs $719$M inference parameters, $37.2$\,ms and $1.57$\,GiB per
image; RAESR-Distilled lowers all three to $502$M, $30.0$\,ms and $1.18$\,GiB. In
Table~\ref{tab:main} only the four smallest baselines carry fewer parameters, and none of them is
close in quality. Only Real-ESRGAN is faster, and it trails on every perceptual metric. RAESR is
$1.6\times$ faster than the fastest one-step diffusion baseline (AdcSR) and $59$--$514\times$
faster than the $20$--$50$-step methods, while TSD-SR, the nearest competitor, is
$3.1\times$ larger and $2.8\times$ slower.

Fig.~\ref{fig:qualitative} shows the qualitative results. The GAN restorers work in pixel space, where
the map is too hard, and thus fall short of the manifold. Their reconstructions are blurred in every
sample. The diffusion baselines pay for the map but still land at the wrong point on the manifold, which shows up as hallucination or as a loss of the original semantics. In the boat
image, RAESR is the only model that reconstructs the canopy in the background and renders the pole and everything around it cleanly. In the second image, it is the only model that preserves
the building in the background. It is also the only model that preserves the real physics of the fox's nose, recreating the fur that sweeps over it in the correct direction.

RAESR-Distilled replaces the frozen DINOv3-L encoder with a distilled frontend and reuses the
released decoder unchanged (App.~\ref{app:distill}). As Table~\ref{tab:main} shows, it largely
preserves the teacher's performance at a smaller size and lower latency. 

\paragraph{Human preference study.}
Our perceptual claim rests on automatic metrics, so we test it directly with human observers. Each trial shows the degraded input and the ground truth above
two anonymous restorations and asks which repair is better, judging both match to the real scene and
how good the image looks. The participants can also choose undecided. The four baselines are Real-HAT-GAN, DiffBIR, SUPIR, and TSD-SR. The pool is $100$ images, $25$
drawn uniformly at random from each benchmark with a fixed seed. Method identity is never shown and left/right assignment is randomised
(App.~\ref{app:human} gives the full protocol). \textbf{Human observers prefer RAESR to every
baseline} (Table~\ref{tab:human}), including a weak win over TSD-SR.

\section{Analysis}
\label{sec:analysis}
This section tests the central hypothesis of the paper, that RAESR's performance comes from the
substrate, which embeds degraded images near the manifold and guides the mapping at every level,
rather than from the decoder. Protocols and full tables are in
Apps.~\ref{app:interp}, \ref{app:channel} and~\ref{app:substrate}.

\subsection{Degraded images are embedded near the manifold}
\label{sec:geometry}

\begin{figure}[t]
\centering
\includegraphics[width=\textwidth]{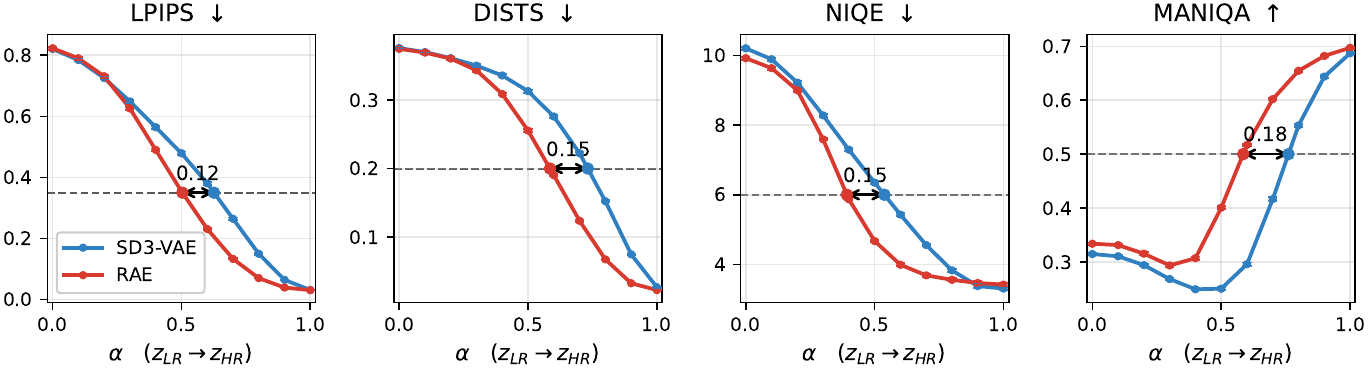}
\caption{Raw metric curves along the latent line
$z_\alpha=(1-\alpha)\,z_{\mathrm{LR}}+\alpha\,z_{\mathrm{HR}}$, decoded at eleven points. Curves are
means over $2{,}000$ images.}
\label{fig:interp}
\end{figure}

For $2{,}000$ images in the LSDIR dataset, we apply the same level of degradation as in the training, embed the degraded and clean images, and decode the latent line $z_\alpha=(1-\alpha)\,z_{\mathrm{LR}}+\alpha\,z_{\mathrm{HR}}$ at eleven evenly distributed points. Two latent spaces are compared: the fused DINOv3-L latent with the RAEv2 decoder finetuned for clean-image reconstruction at $512^2$ and the SD3 VAE. We evaluate four metrics, LPIPS, DISTS, NIQE and MANIQA, on the decoded images at those eleven points along the latent line and then averaged them across all images. The comparison is endpoint-matched. Fig.~\ref{fig:interp} shows a consistent picture. On all four metrics, the line in the fused latent recovers the statistics of the clean image much faster than in the VAE latent, and the two curves separate most widely in
the middle. The degraded image is therefore encoded relatively closer both to the manifold and to the embedding of the faithful high-resolution image. The fused latent embeds the degraded image near the manifold, so a short move along the return path already restores naturalness and sharpness. Threshold robustness, the SD2.1 replication and the control experiments are in App.~\ref{app:interp}.

\begin{figure}[t]
\centering
\includegraphics[width=\textwidth]{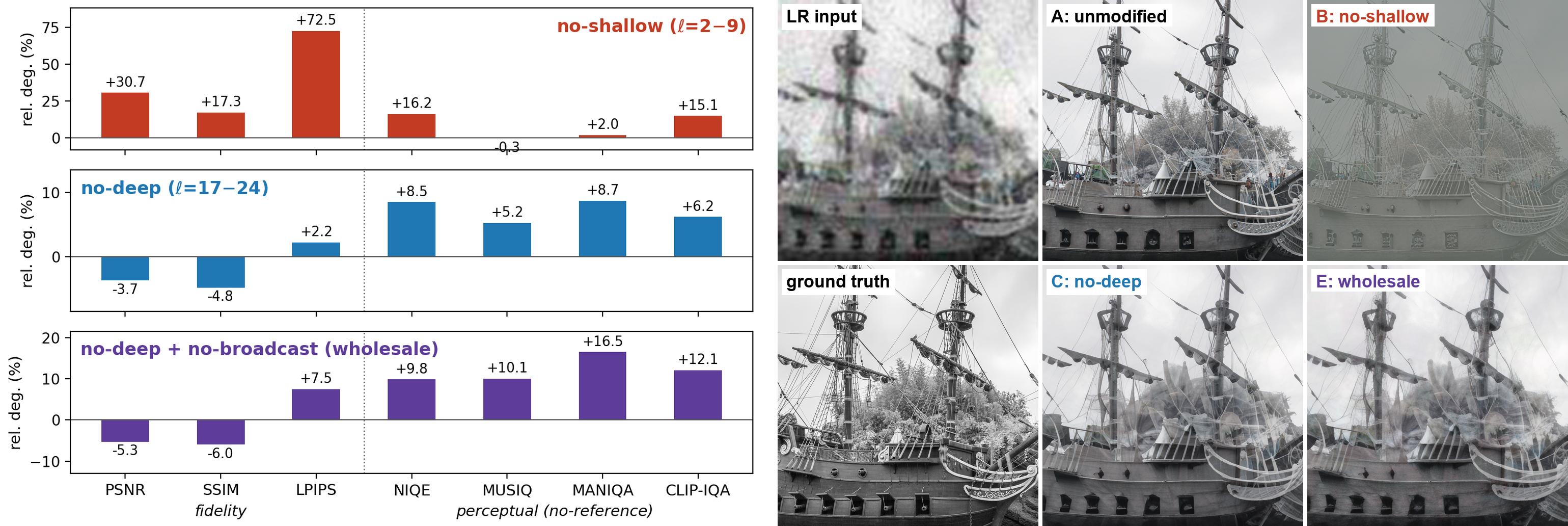}
\caption{ \emph{Left:} relative change in each metric
when one eight-layer band of the average is removed;
negative means the metric improves. \emph{Right:} the same interventions on one LSDIR image.}
\label{fig:intervention}
\end{figure}

\subsection{The decoder relies on every level of the substrate}
\label{sec:decoder-reads}
Each level of information in the substrate is essential to guiding the map to the correct
embedding on the manifold, and the decoder relies on all of them. With every weight frozen, we replace one band of the average with its training-set mean tokens, and re-score the
released model: the shallowest eight layers, the deepest eight, and additionally the broadcast term (Fig.~\ref{fig:intervention}). Removing the shallow band severely deteriorates the fidelity metrics
, while the perceptual metrics deteriorate only slightly.
Removing the deep band does the opposite. Removing the broadcast term as well widens both effects. We also randomly remove 8 layers and the deterioration shows no trend (App.~\ref{app:channel}). The sample images on the right show two distinct failure modes. Without the shallow band the output loses fidelity to the input. The color
balance and the contrast shift, and structure is synthesised without reference to the observed
pixels. Without the
deep semantics-rich band, the decoder cannot determine
what to synthesise, and it selects the safest option, a plausible generic texture.

\begin{figure}[t]
\centering
\includegraphics[width=\textwidth]{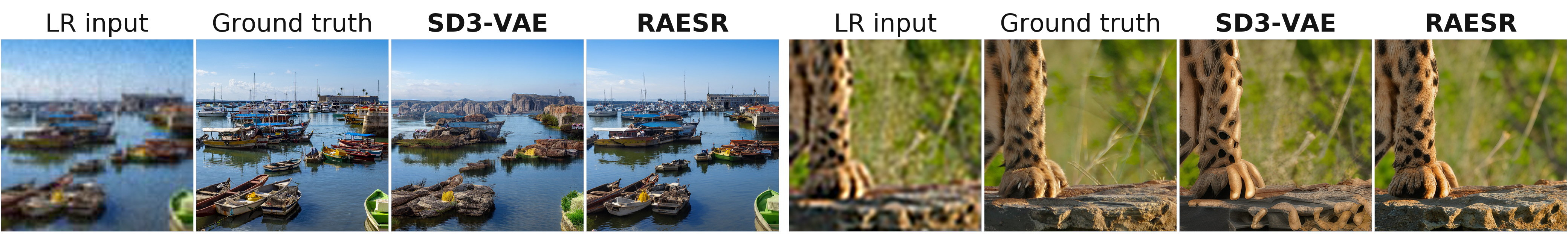}\\[5pt]
{\footnotesize\setlength{\tabcolsep}{3.4pt}
\begin{tabular}{lccccccccc}
\toprule
Substrate & PSNR$\uparrow$ & SSIM$\uparrow$ & LPIPS$\downarrow$ & DISTS$\downarrow$ & NIQE$\downarrow$ & MUSIQ$\uparrow$ & MANIQA$\uparrow$ & TOPIQ$\uparrow$ & CLIP$\uparrow$ \\
\midrule
DINOv3-L & \textbf{23.72} & \textbf{0.6265} & \textbf{0.3012} & \textbf{0.1971} & \textbf{4.320} & \textbf{70.45} & \textbf{0.6717} & \textbf{0.6820} & \textbf{0.7059} \\
SD3 VAE & 23.61 & 0.6033 & 0.3734 & 0.2495 & 4.344 & 69.76 & 0.6300 & 0.6632 & 0.7011 \\
\bottomrule
\end{tabular}}
\caption{\textbf{Swapping the substrate, keeping everything else.} \textbf{Bold} marks the
better score of the two. Two further examples in
Fig.~\ref{fig:substrate-extra}.}
\label{fig:substrate}
\end{figure}

\subsection{The semantics come from the substrate, not the decoder}
\label{sec:substrate}
The remaining attribution question is whether the result belongs to the representation or to the
$415$M decoder and its recipe. It is reasonable to suspect that the decoder could ignore the semantics in the latent and learn its own from the pixel record. We
keep everything and only replace the frozen DINOv3-L with the frozen SD3 VAE \citep{sd3}. The decoder, its initialisation and the three-stage recipe are unchanged up to minor deviations (App.~\ref{app:substrate}). \textbf{The VAE substrate loses on all nine metrics} (table in
Fig.~\ref{fig:substrate}). Under heavy degradation the model either
paints confident, sharp, \emph{wrong} content or does not produce a clean, natural image at all,
because the degraded image is encoded too far from the manifold and the substrate provides too
little direct semantic guidance.

\section{Conclusion and Limitations}
\label{sec:conclusion}
Real-world SR is the task of mapping a degraded image back onto the manifold of clean, sharp and
natural images, and to the point on it that is faithful to the input. Which latent space the
mapping is performed in therefore decides how hard the task is. A suitable substrate embeds
degraded images near the manifold, so the map back is short, and carries information at every
level, from the pixel record to semantics, so the map is guided to the faithful point. The fused
layers of a frozen DINOv3-L are such a space. With only one decoder pass, our RAESR attains the best fidelity--perception and efficiency--perception
trade-offs among the twelve published methods we evaluate. Human observers prefer its restorations to those of four competitive models we evaluated. We therefore argue that the choice of substrate
deserves more attention than the SR literature currently gives.

\paragraph{Limitations.}
RAESR has two limitations. Its restorations are produced in one pass with no mechanism for
trading compute for quality, so unlike the multi-step methods it cannot spend more steps on a
hard image. Our claim about what makes a substrate suitable is supported on one encoder family.
We characterise the fused layers of DINOv3-L and test a single alternative substrate
(\S\ref{sec:substrate}), leaving open how other representation encoders behave.

\bibliography{refs}
\bibliographystyle{plainnat}

\newpage
\beginappendix
\section{Human preference study}
\label{app:human}

\paragraph{Protocol.} Each trial shows the degraded input and the ground truth, labelled in plain
words, above two anonymous restorations, and asks which repair is better, judging together whether
the output matches the real scene, with nothing invented or changed, and whether it looks sharp and
natural. A third option records a tie. Showing the ground truth is deliberate: it makes the study
measure faithfulness and quality together, which is the claim of the paper, rather than realism
alone. The pool is $100$ images, $25$ drawn uniformly at random from each benchmark's evaluation
set with a fixed seed and no author selection at any stage, restored by the same official
checkpoints and defaults as Table~\ref{tab:main} and served at native $512{\times}512$ with
tap-to-zoom. Method identity is never shown, left/right assignment is randomised per trial, trial
order is shuffled per participant, and each participant judges $25$ distinct images so no scene is
seen twice within a session.

\paragraph{Quality control.} Three catch trials per participant, mixed invisibly into the
sequence, compare RAESR against the bicubically upsampled input; a participant failing more than
one is excluded entirely, as is one whose median response time falls under $3$\,s. Both rules were
fixed before collection. The preference rate is RAESR votes over the sum of RAESR and baseline
votes, with ties excluded and reported in Table~\ref{tab:human}.

\section{Evaluation details and per-benchmark results}
\label{app:perset}

This appendix documents how Table~\ref{tab:main} was produced. It gives the evaluation protocol
(App.~\ref{app:evalproto}), the configurations and training signals of the baselines
(App.~\ref{app:baselines}), the per-benchmark results behind the four-benchmark average
(App.~\ref{app:perbench}), the fused perceptual score plotted in Fig.~\ref{fig:tradeoff}
(App.~\ref{app:fused}), and the effect of the channel-mean alignment used at inference
(App.~\ref{app:meanalign}).

\subsection{Evaluation protocol}
\label{app:evalproto}

PSNR and SSIM are computed on the luminance (Y) channel of the YCbCr colour space, following
common SR practice, and MANIQA uses the \texttt{maniqa-pipal} checkpoint. The metric computation is
deterministic to machine precision, so a scored checkpoint reproduces its numbers exactly and single
entries carry no evaluation error bar. 

\subsection{Baseline configurations and training signals}
\label{app:baselines}

Every baseline row of Table~\ref{tab:main} uses the method's official checkpoint at its official
defaults, and no unofficial checkpoint was substituted for any row. SUPIR runs its full official
pipeline, LLaVA-13B captioner included, at its official $1024$px internal size, with the output
bicubically reduced to the shared $512$. DiffBIR is reported in its published v2 configuration, which takes no text input.

\paragraph{Training signals.} Table~\ref{tab:losses} summarises what enters each baseline's \emph{training} loss, compiled from
the official papers and, where released, official training code (SUPIR releases no training code;
its paper states no quality-model term, and its negative-quality \emph{captions} are a text-side
signal). No baseline trains on a no-reference IQA model; every NIQE/MUSIQ/MANIQA/CLIP-IQA number in
this paper is therefore held out from training for every row, ours included
(App.~\ref{app:training}).

\begin{table}[!ht]
\centering
\renewcommand{\arraystretch}{0.92}
\footnotesize
\setlength{\tabcolsep}{4pt}
\caption{\textbf{Training signals of the baselines} (coarse; official paper/code). Ref.\
perc.\ = reference-based perceptual loss; NR-IQA = a no-reference quality model as a training
signal.}
\label{tab:losses}
\begin{tabular}{lllll}
\toprule
Method & Reconstruction & Ref.\ perceptual & Adversarial & NR-IQA \\
\midrule
Real-ESRGAN & L1 & VGG feature & GAN & none \\
StableSR & latent $\epsilon$-MSE; CFW: L1 & CFW: LPIPS & CFW: GAN & none \\
ResShift (4-step) & latent MSE & LPIPS & --- & none \\
OSEDiff & pixel MSE $+$ VSD & LPIPS & --- & none \\
DiffBIR & MSE / latent $\epsilon$-MSE & --- & --- & none \\
SUPIR & (training code unreleased) & --- & --- & none stated \\
VOSR-1.4B-ms & velocity MSE & --- & --- & none \\
VOSR-1.4B-os & teacher-velocity MSE $+$ consistency & --- & --- & none \\
\midrule
Real-HAT-GAN & L1 & VGG feature & GAN & none \\
TSD-SR & MSE $+$ target score distillation & LPIPS & --- & none \\
DiT4SR & flow-matching MSE & --- & --- & none \\
AdcSR & latent L1 distillation & --- & GAN (UNet feat.) & none \\
\midrule
\textbf{RAESR} & L1 & LPIPS & GAN & \textbf{none} \\
\bottomrule
\end{tabular}
\end{table}

\subsection{Per-benchmark results}
\label{app:perbench}

Table~\ref{tab:main} reports the average over the four benchmarks.
Tables~\ref{tab:perset-lsdir}--\ref{tab:perset-div2k} break it down by benchmark, with the same
protocol and the same bold and underline marking. Excluding LSDIR, the one in-distribution
benchmark, RAESR still leads on NIQE, DISTS and MANIQA and is second on MUSIQ and TOPIQ.

\begin{table}[t]
\centering
\renewcommand{\arraystretch}{0.92}
\footnotesize
\setlength{\tabcolsep}{2.8pt}
\caption{\textbf{Per-benchmark results: LSDIR} (250 images). Marks and protocol as in Table~\ref{tab:main}.}
\label{tab:perset-lsdir}
\resizebox{\textwidth}{!}{%
\begin{tabular}{lcccccccccc}
\toprule
Method & NFE & PSNR$\uparrow$ & SSIM$\uparrow$ & LPIPS$\downarrow$ & DISTS$\downarrow$ & NIQE$\downarrow$ & MUSIQ$\uparrow$ & MANIQA$\uparrow$ & TOPIQ$\uparrow$ & CLIP$\uparrow$ \\
\midrule
\textbf{RAESR} & 1 & 19.77 & 0.4354 & \underline{0.3485} & \underline{0.1923} & \textbf{3.312} & 73.58 & \underline{0.6987} & \underline{0.7192} & 0.7108 \\
\textbf{RAESR}$^{\rm D}$ & 1 & 19.92 & 0.4409 & 0.3629 & 0.2098 & \underline{3.416} & 73.13 & 0.6859 & 0.7062 & 0.7039 \\
\midrule
\multicolumn{11}{l}{\emph{Adversarial restorers}} \\
Real-ESRGAN & 1 & 19.95 & \underline{0.4694} & 0.4135 & 0.2526 & 3.942 & 66.22 & 0.5663 & 0.5997 & 0.5899 \\
Real-HAT-GAN & 1 & \underline{20.55} & \textbf{0.4791} & 0.4289 & 0.2769 & 4.345 & 59.37 & 0.5433 & 0.4852 & 0.4570 \\
\midrule
\multicolumn{11}{l}{\emph{Diffusion trained from scratch}} \\
ResShift & 4 & \textbf{20.55} & 0.4642 & 0.4252 & 0.2551 & 6.300 & 59.57 & 0.5113 & 0.4941 & 0.5419 \\
VOSR-1.4B-ms & 25 & 19.53 & 0.4425 & 0.3612 & \textbf{0.1854} & 4.527 & \underline{74.28} & \textbf{0.7001} & 0.6806 & 0.6197 \\
VOSR-1.4B-os & 1 & 19.45 & 0.4350 & 0.3695 & 0.1948 & 4.081 & 73.83 & 0.6637 & 0.6548 & 0.5811 \\
\midrule
\multicolumn{11}{l}{\emph{Diffusion on a T2I basis}} \\
AdcSR & 1 & 19.74 & 0.4316 & 0.3666 & 0.2104 & 3.521 & 70.23 & 0.6122 & 0.6753 & 0.6711 \\
StableSR & 20 & 19.40 & 0.4064 & 0.4082 & 0.2349 & 4.002 & 67.54 & 0.6167 & 0.6333 & 0.6759 \\
DiffBIR & 50 & 19.71 & 0.4190 & 0.4094 & 0.2134 & 4.447 & 71.89 & 0.6608 & \textbf{0.7275} & \textbf{0.7332} \\
OSEDiff & 1 & 20.02 & 0.4534 & 0.3706 & 0.2227 & 3.886 & 69.86 & 0.6183 & 0.6378 & 0.6500 \\
TSD-SR & 1 & 19.04 & 0.4170 & \textbf{0.3329} & 0.2019 & 3.584 & \textbf{74.29} & 0.6353 & 0.6995 & \underline{0.7306} \\
DiT4SR & 40 & 18.18 & 0.3993 & 0.4160 & 0.2129 & 4.309 & 72.93 & 0.6846 & 0.6969 & 0.6917 \\
SUPIR & 50 & 19.07 & 0.3745 & 0.4697 & 0.2605 & 5.595 & 64.79 & 0.6068 & 0.6531 & 0.7079 \\
\bottomrule
\end{tabular}}
\end{table}

\begin{table}[t]
\centering
\renewcommand{\arraystretch}{0.92}
\footnotesize
\setlength{\tabcolsep}{2.8pt}
\caption{\textbf{Per-benchmark results: RealSR} (100 images). Marks and protocol as in Table~\ref{tab:main}.}
\label{tab:perset-realsr}
\resizebox{\textwidth}{!}{%
\begin{tabular}{lcccccccccc}
\toprule
Method & NFE & PSNR$\uparrow$ & SSIM$\uparrow$ & LPIPS$\downarrow$ & DISTS$\downarrow$ & NIQE$\downarrow$ & MUSIQ$\uparrow$ & MANIQA$\uparrow$ & TOPIQ$\uparrow$ & CLIP$\uparrow$ \\
\midrule
\textbf{RAESR} & 1 & 25.00 & 0.7195 & \underline{0.2703} & \underline{0.2012} & \textbf{4.587} & 70.49 & \textbf{0.6757} & 0.6705 & 0.6841 \\
\textbf{RAESR}$^{\rm D}$ & 1 & 23.45 & 0.7045 & 0.2864 & 0.2141 & \underline{4.856} & \underline{70.77} & \underline{0.6748} & 0.6718 & 0.6852 \\
\midrule
\multicolumn{11}{l}{\emph{Adversarial restorers}} \\
Real-ESRGAN & 1 & 25.00 & \underline{0.7473} & 0.2753 & 0.2076 & 5.807 & 60.43 & 0.5500 & 0.5163 & 0.4581 \\
Real-HAT-GAN & 1 & \textbf{26.55} & \textbf{0.7655} & \textbf{0.2480} & \textbf{0.1969} & 6.348 & 57.11 & 0.5513 & 0.4458 & 0.3737 \\
\midrule
\multicolumn{11}{l}{\emph{Diffusion trained from scratch}} \\
ResShift & 4 & \underline{25.02} & 0.7253 & 0.3255 & 0.2461 & 7.915 & 57.70 & 0.5160 & 0.4583 & 0.5365 \\
VOSR-1.4B-ms & 25 & 24.70 & 0.7047 & 0.2853 & 0.2114 & 5.902 & 67.50 & 0.6142 & 0.5900 & 0.5118 \\
VOSR-1.4B-os & 1 & 24.60 & 0.6990 & 0.2782 & 0.2064 & 5.312 & 70.71 & 0.6483 & 0.6410 & 0.5634 \\
\midrule
\multicolumn{11}{l}{\emph{Diffusion on a T2I basis}} \\
AdcSR & 1 & 24.98 & 0.7175 & 0.2917 & 0.2145 & 5.343 & 69.89 & 0.6339 & \textbf{0.6798} & 0.6709 \\
StableSR & 20 & 24.20 & 0.6984 & 0.3039 & 0.2166 & 5.808 & 65.23 & 0.6214 & 0.5670 & 0.6150 \\
DiffBIR & 50 & 24.16 & 0.6435 & 0.3667 & 0.2411 & 5.658 & 69.50 & 0.6475 & \underline{0.6733} & \underline{0.7000} \\
OSEDiff & 1 & 24.63 & 0.7193 & 0.2968 & 0.2144 & 5.705 & 69.09 & 0.6321 & 0.6257 & 0.6702 \\
TSD-SR & 1 & 24.24 & 0.7025 & 0.2776 & 0.2127 & 5.110 & \textbf{71.20} & 0.6341 & 0.6598 & \textbf{0.7172} \\
DiT4SR & 40 & 22.85 & 0.6432 & 0.3271 & 0.2271 & 6.255 & 67.98 & 0.6640 & 0.6015 & 0.6361 \\
SUPIR & 50 & 24.56 & 0.6611 & 0.3741 & 0.2491 & 7.712 & 59.44 & 0.5775 & 0.5431 & 0.6436 \\
\bottomrule
\end{tabular}}
\end{table}

\begin{table}[t]
\centering
\renewcommand{\arraystretch}{0.92}
\footnotesize
\setlength{\tabcolsep}{2.8pt}
\caption{\textbf{Per-benchmark results: DRealSR} (93 images). Marks and protocol as in Table~\ref{tab:main}.}
\label{tab:perset-drealsr}
\resizebox{\textwidth}{!}{%
\begin{tabular}{lcccccccccc}
\toprule
Method & NFE & PSNR$\uparrow$ & SSIM$\uparrow$ & LPIPS$\downarrow$ & DISTS$\downarrow$ & NIQE$\downarrow$ & MUSIQ$\uparrow$ & MANIQA$\uparrow$ & TOPIQ$\uparrow$ & CLIP$\uparrow$ \\
\midrule
\textbf{RAESR} & 1 & 27.85 & 0.7613 & 0.3030 & 0.2120 & \textbf{5.348} & \textbf{67.41} & \textbf{0.6449} & 0.6517 & 0.7091 \\
\textbf{RAESR}$^{\rm D}$ & 1 & 26.52 & 0.7567 & 0.3102 & 0.2207 & \underline{5.531} & \underline{67.40} & \underline{0.6405} & 0.6477 & 0.7132 \\
\midrule
\multicolumn{11}{l}{\emph{Adversarial restorers}} \\
Real-ESRGAN & 1 & \underline{28.61} & \underline{0.8051} & \underline{0.2819} & \underline{0.2089} & 6.698 & 54.27 & 0.4903 & 0.4624 & 0.4516 \\
Real-HAT-GAN & 1 & \textbf{30.02} & \textbf{0.8324} & \textbf{0.2605} & \textbf{0.2063} & 7.294 & 50.56 & 0.4915 & 0.4053 & 0.3916 \\
\midrule
\multicolumn{11}{l}{\emph{Diffusion trained from scratch}} \\
ResShift & 4 & 27.11 & 0.7448 & 0.3860 & 0.2619 & 8.739 & 50.49 & 0.4617 & 0.4477 & 0.5398 \\
VOSR-1.4B-ms & 25 & 27.50 & 0.7324 & 0.3754 & 0.2490 & 7.373 & 57.26 & 0.5277 & 0.5103 & 0.4822 \\
VOSR-1.4B-os & 1 & 27.59 & 0.7356 & 0.3291 & 0.2311 & 6.216 & 65.72 & 0.6061 & 0.6016 & 0.5871 \\
\midrule
\multicolumn{11}{l}{\emph{Diffusion on a T2I basis}} \\
AdcSR & 1 & 28.10 & 0.7726 & 0.3046 & 0.2200 & 6.450 & 66.27 & 0.5916 & \underline{0.6528} & 0.7052 \\
StableSR & 20 & 28.08 & 0.7509 & 0.3307 & 0.2295 & 6.592 & 58.68 & 0.5551 & 0.5378 & 0.6238 \\
DiffBIR & 50 & 25.55 & 0.6071 & 0.4772 & 0.2889 & 6.284 & 66.82 & 0.6274 & \textbf{0.6796} & \underline{0.7136} \\
OSEDiff & 1 & 27.91 & 0.7833 & 0.2967 & 0.2164 & 6.444 & 64.70 & 0.5903 & 0.6003 & 0.6970 \\
TSD-SR & 1 & 27.77 & 0.7560 & 0.2966 & 0.2136 & 5.910 & 66.61 & 0.5873 & 0.6181 & \textbf{0.7344} \\
DiT4SR & 40 & 25.58 & 0.6679 & 0.3819 & 0.2506 & 6.820 & 65.07 & 0.6338 & 0.5906 & 0.6697 \\
SUPIR & 50 & 26.82 & 0.6705 & 0.4303 & 0.2770 & 9.029 & 54.63 & 0.5332 & 0.5080 & 0.6318 \\
\bottomrule
\end{tabular}}
\end{table}

\begin{table}[t]
\centering
\renewcommand{\arraystretch}{0.92}
\footnotesize
\setlength{\tabcolsep}{2.8pt}
\caption{\textbf{Per-benchmark results: DIV2K-Val} (3000 images). Marks and protocol as in Table~\ref{tab:main}.}
\label{tab:perset-div2k}
\resizebox{\textwidth}{!}{%
\begin{tabular}{lcccccccccc}
\toprule
Method & NFE & PSNR$\uparrow$ & SSIM$\uparrow$ & LPIPS$\downarrow$ & DISTS$\downarrow$ & NIQE$\downarrow$ & MUSIQ$\uparrow$ & MANIQA$\uparrow$ & TOPIQ$\uparrow$ & CLIP$\uparrow$ \\
\midrule
\textbf{RAESR} & 1 & 23.71 & 0.5915 & \underline{0.2810} & \underline{0.1827} & \textbf{4.027} & 70.55 & \textbf{0.6658} & \underline{0.6845} & 0.7158 \\
\textbf{RAESR}$^{\rm D}$ & 1 & 23.66 & 0.5958 & 0.2913 & 0.1943 & \underline{4.136} & 70.46 & \underline{0.6595} & 0.6764 & 0.7108 \\
\midrule
\multicolumn{11}{l}{\emph{Adversarial restorers}} \\
Real-ESRGAN & 1 & 24.29 & \underline{0.6372} & 0.3112 & 0.2141 & 4.679 & 61.06 & 0.5485 & 0.5297 & 0.5277 \\
Real-HAT-GAN & 1 & \textbf{25.10} & \textbf{0.6547} & 0.3066 & 0.2279 & 5.361 & 55.20 & 0.5378 & 0.4370 & 0.4124 \\
\midrule
\multicolumn{11}{l}{\emph{Diffusion trained from scratch}} \\
ResShift & 4 & \underline{24.60} & 0.6240 & 0.3064 & 0.2131 & 6.932 & 59.01 & 0.5279 & 0.5072 & 0.5717 \\
VOSR-1.4B-ms & 25 & 23.73 & 0.6036 & 0.3038 & 0.1880 & 5.471 & 70.55 & 0.6542 & 0.6364 & 0.6022 \\
VOSR-1.4B-os & 1 & 23.41 & 0.5914 & 0.3093 & 0.1904 & 5.029 & \underline{70.59} & 0.6412 & 0.6266 & 0.5903 \\
\midrule
\multicolumn{11}{l}{\emph{Diffusion on a T2I basis}} \\
AdcSR & 1 & 23.74 & 0.6017 & 0.2853 & 0.1899 & 4.358 & 67.99 & 0.6073 & 0.6526 & 0.6763 \\
StableSR & 20 & 23.28 & 0.5733 & 0.3104 & 0.2043 & 4.743 & 65.83 & 0.6172 & 0.5956 & 0.6758 \\
DiffBIR & 50 & 23.15 & 0.5438 & 0.3670 & 0.2214 & 5.035 & 69.91 & 0.6434 & \textbf{0.7038} & \underline{0.7304} \\
OSEDiff & 1 & 23.72 & 0.6109 & 0.2942 & 0.1975 & 4.710 & 67.97 & 0.6132 & 0.6188 & 0.6683 \\
TSD-SR & 1 & 23.02 & 0.5808 & \textbf{0.2673} & \textbf{0.1821} & 4.326 & \textbf{71.69} & 0.6192 & 0.6679 & \textbf{0.7416} \\
DiT4SR & 40 & 21.70 & 0.5441 & 0.3468 & 0.2119 & 4.950 & 68.47 & 0.6467 & 0.6301 & 0.6692 \\
SUPIR & 50 & 23.15 & 0.5432 & 0.3631 & 0.2257 & 6.328 & 62.58 & 0.5904 & 0.6126 & 0.7030 \\
\bottomrule
\end{tabular}}
\end{table}

\subsection{The fused perceptual score of Figure~\ref{fig:tradeoff}}
\label{app:fused}

The vertical axis of Figure~\ref{fig:tradeoff} fuses the six perceptual columns of
Table~\ref{tab:main} (DISTS, NIQE, MUSIQ, MANIQA, TOPIQ, CLIP-IQA) into one number per method.
DISTS and NIQE are negated so that higher is better on all six. Each metric is then min--max
normalised over the $14$ rows,
\begin{equation}
s_{i,m} \;=\; \frac{v_{i,m} - \min_j v_{j,m}}{\max_j v_{j,m} - \min_j v_{j,m}},
\end{equation}
and the fused score of method $i$ is the unweighted mean $\frac{1}{6}\sum_m s_{i,m}$. The best
method on a metric receives $1$ and the worst $0$, so the score is bounded in $[0,1]$ and every
metric contributes equally.

Min--max normalisation rescales each metric by its \emph{observed} spread, so a metric on which
all methods nearly tie would have its noise stretched to the full unit interval and be
overweighted. That failure mode does not arise here: on every one of the six metrics the spread
across the $14$ methods is $23$--$58\%$ of the metric's mean value. The ranking is also invariant
to the normalisation: replacing min--max by per-metric $z$-scores reproduces the ordering of all
$14$ methods exactly, and replacing it by the mean of per-metric ranks keeps the top position, the
entire bottom half, and changes only adjacent mid-field pairs whose raw values are close (e.g.\
MUSIQ $70.95$ against $70.51$). 

\subsection{Channel-mean alignment}
\label{app:meanalign}

The inference step of \S\ref{sec:pipeline} aligns each output channel's mean to the corresponding
channel mean of the bicubically upsampled input. Averaged over the four benchmarks it adds $0.36$\,dB
PSNR to the teacher and $0.99$\,dB to the student, and costs nothing measurable: on the teacher MUSIQ
rises $0.06$ and no other metric moves by as much as $0.001$ in either direction. 

\section{The frozen encoder: full geometry}
\label{app:geometry}

This appendix backs the analysis of the frozen encoder in \S\ref{sec:contains}. All measurements are
on DINOv3-L/16. Layer $\ell$ denotes the normalised patch-token output of block $\ell$ for
$\ell = 1,\dots,24$, and CLS and register tokens are excluded. The latent of Eq.~\ref{eq:agg}
averages layers $2$ to $24$ and adds the spatial mean of layer $24$, so layer $1$ is not in it.

\subsection{Layer disentanglement and redundancy}
\label{app:disentangle}
\begin{figure}[t]
\centering
\includegraphics[clip, trim=389 0 0 0, width=0.78\textwidth]{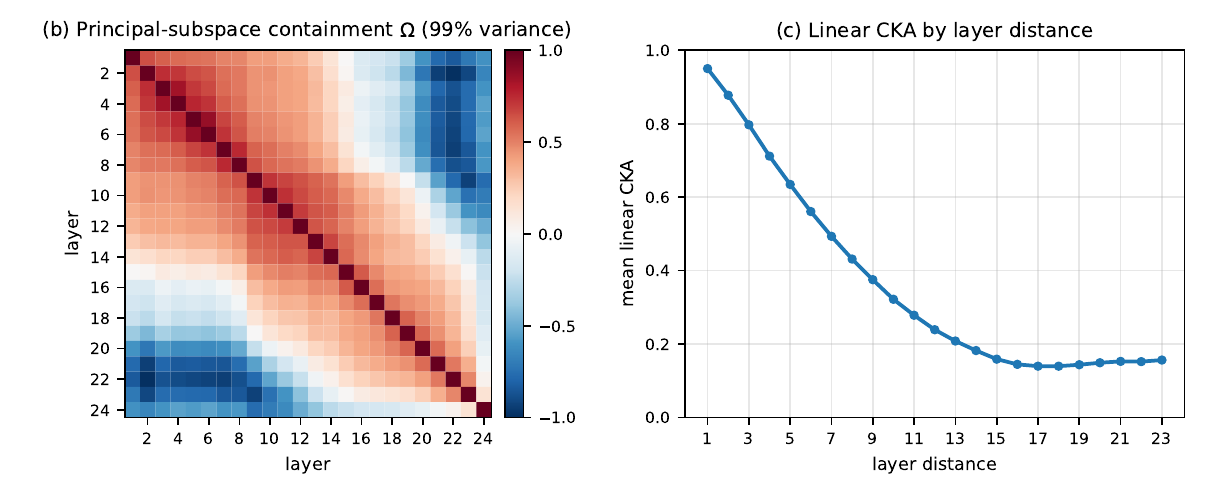}
\caption{Layer-pair geometry on training-degraded inputs: chance-corrected, variance-matched
containment $\Omega$ (left) and linear CKA against layer distance (right).}
\label{fig:geom-banded}
\end{figure}

Averaging twenty-three $1{,}024$-dimensional layers costs little, for the geometric reasons
summarised in \S\ref{sec:contains} and measured here. An
average cannot be lossless, and it would lose the most if the layers wrote into the same
directions. They do not. On training-degraded inputs, the distribution the model sees, we measure
the overlap of every pair of layers' variance-matched principal subspaces, corrected for chance,
and the linear CKA between their token-similarity structures. Let $U_i \in \mathbb{R}^{1024\times
p}$ hold the leading principal directions of layer $i$'s tokens that explain $99\%$ of their
variance, $U_j \in \mathbb{R}^{1024\times q}$ the same for layer $j$, and $X_i, X_j \in
\mathbb{R}^{n\times 1024}$ the centred token matrices of the two layers over the same $n$ tokens.
The chance-corrected containment and the linear CKA \citep{cka} are
\begin{equation}
\Omega_{ij} = \frac{\lVert U_i^{\top} U_j \rVert_F^{2} - pq/1024}{\min(p,q) - pq/1024},
\qquad
\mathrm{CKA}_{ij} = \frac{\lVert X_j^{\top} X_i \rVert_F^{2}}{\lVert X_i^{\top} X_i \rVert_F
\,\lVert X_j^{\top} X_j \rVert_F}.
\label{eq:containment}
\end{equation}
Here $\Omega_{ij} = 1$ means that one subspace is contained in the other, $0$ means the overlap two
random subspaces would have, and a negative value means less than chance; CKA is the cosine
similarity between the two layers' token-by-token similarity matrices, invariant to any rotation
and rescaling of the feature axes. Both statistics tell the same story
(Fig.~\ref{fig:geom-banded}).
Distant layers are disentangled. Subspace overlap falls with layer distance, drops below what
random subspaces would share eleven layers apart, and reaches $-0.60$ between the first and last
layers, while CKA flattens at $0.14$--$0.16$ beyond fifteen layers apart. Neighbouring layers are redundant: adjacent overlap is $0.64$ and adjacent CKA $0.95$, so each level is held in several nearby copies, stored in different forms. The recovery experiment of Fig.~\ref{fig:geom-depth}b,c measures the same redundancy directly. Control N3 predicts each layer linearly from the single other layer that predicts it best, which is always an adjacent layer, and reaches $R^2$ of $0.865$ to $0.988$ on clean inputs and $0.783$ to $0.998$ on degraded inputs (App.~\ref{app:recovery}). Disentanglement means that the average superposes
distinct levels rather than collapsing them onto one another, and redundancy means that what the
sum costs one layer, its neighbours still carry.

\subsection{Recovery of each layer from the fused latent}
\label{app:recovery}
\label{app:degraded}

This experiment backs Fig.~\ref{fig:geom-depth}b,c. We take $4{,}096$ LSDIR images and eight fixed
patch positions in each. For every image we compute the fused latent $z$ of Eq.~\ref{eq:agg} and the
tokens $h_\ell$ of all $24$ layers at those positions, once on the clean image and once on its version
degraded by the training degradation pipeline. For each target layer $\ell$ we fit a multivariate ridge
regression from $z$ to $h_\ell$. The images are split by image into training, validation and test sets
($2{,}867/614/615$), so the eight tokens of an image never cross sets. The ridge strength is selected
on the validation set from $\{10^{-6},\dots,10\}$ times the mean diagonal of the centred training
$X^{\top}X$, and we report the held-out
\begin{equation}
  R^2 \;=\; 1 - \frac{\sum_{p}\lVert \hat{h}_\ell(p) - h_\ell(p) \rVert^2}
                     {\sum_{p}\lVert h_\ell(p) - \bar{h}_\ell \rVert^2},
  \label{eq:r2}
\end{equation}
where the sums run over all test tokens $p$, $\hat{h}_\ell$ is the ridge prediction and $\bar{h}_\ell$
is the mean training token.

On clean inputs $R^2$ ranges from $0.695$ to $0.866$ with a mean of $0.820$. The lowest value is at
layer $1$, which the latent does not contain. On degraded inputs $R^2$ ranges from $0.646$ to $0.893$.
Degradation makes the shallow layers easier to recover ($0.695 \rightarrow 0.893$ at layer $1$) and the
deep layers harder ($0.839 \rightarrow 0.646$ at layer $24$). Since the map is linear, these values are
lower bounds on what the decoder can extract.

Fig.~\ref{fig:geom-depth}b,c also plots three controls, fitted and scored in the same way. N1 feeds the
map the latent of a different image, N3 predicts each layer from the single other layer that predicts
it best, and N4 predicts the training-set mean token. N1 stays near zero on clean inputs (at most
$0.006$) and below $0.13$ on degraded inputs, and N4 gives $R^2 \approx 0$, so the recovery is specific
to each image. The best single predictor of a layer is always an adjacent layer, with $R^2$ from $0.865$ to $0.988$ on clean inputs and from $0.783$ to $0.998$ on degraded inputs, and N3 lies above the fused latent at every layer (Fig.~\ref{fig:geom-depth}b,c). A single neighbour holds nearly all of a layer's content, so what the average dilutes in one layer is still carried by its neighbours. This is the neighbouring-layer redundancy of App.~\ref{app:disentangle}, and it is why averaging loses little.

The lower recovery of the deep layers is consistent with their dimensionality. The number of principal
components that explain $95\%$ of a layer's token variance rises from $9$ at layer $1$ to $404$ at
layer $24$. A target that spreads its variance over many directions overlaps more with the other
layers inside the shared $1{,}024$-dimensional average and is harder to recover linearly. Degradation
widens this difference. It lowers the dimension of the shallow layers ($28 \rightarrow 19$ components
at layer $7$) and raises it for the deep layers ($285 \rightarrow 326$ at layer $19$), and recovery
moves the same way, up for the shallow layers and down for the deep ones.

\section{Latent-line geometry: protocol, thresholds and replications}
\label{app:interp}

\paragraph{Protocol.} The suite behind \S\ref{sec:geometry} uses the first $2{,}000$ images of
a fixed-permutation LSDIR manifest (seed $20260917$; manifest order everywhere, no curation),
resized and centre-cropped to $512^2$. The degraded version is the training profile: two-order
Real-ESRGAN degradation at native $128^2$, bicubically upsampled to $512^2$. Latent spaces:
\emph{rae\_512\_v2}, the RAEv2 ViT-XL decoder initialised from the official release and
finetuned for clean-image reconstruction at $512^2$ (reconstruction losses only; no SR objective
and no degraded inputs), and the SD3 VAE in deterministic posterior mode; the SD2.1 VAE serves
as a replication. No SR-trained checkpoint touches any decode. The line
$z_\alpha=(1-\alpha)z_{\mathrm{LR}}+\alpha z_{\mathrm{HR}}$ is decoded at
$\alpha\in\{0,0.1,\dots,1\}$; decodes are scored with the same metric implementations as the
benchmark evaluation, and means carry image-level bootstrap CIs ($2{,}000$ resamples). 

\begin{table}[h]
\centering\small
\caption{Path fraction $\alpha$ at which the mean raw curve first crosses each threshold
(smaller: the decoded line turns natural, or reaches the ground truth, earlier). The RAE column
is shared by both studies, which use the same RAE rows.}
\label{tab:interp-thresholds}
\begin{tabular}{lccc}
\toprule
Threshold & RAE & SD3-VAE & SD2.1-VAE \\
\midrule
NIQE $\le 5$    & $0.47$ & $0.65$ & $0.68$ \\
NIQE $\le 6$    & $0.39$ & $0.54$ & $0.54$ \\
NIQE $\le 7$    & $0.33$ & $0.43$ & $0.43$ \\
LPIPS $\le 0.25$ & $0.59$ & $0.71$ & $0.73$ \\
LPIPS $\le 0.35$ & $0.50$ & $0.63$ & $0.63$ \\
LPIPS $\le 0.50$ & $0.39$ & $0.48$ & $0.46$ \\
\bottomrule
\end{tabular}
\end{table}

\begin{figure}[t]
\centering
\includegraphics[width=\textwidth]{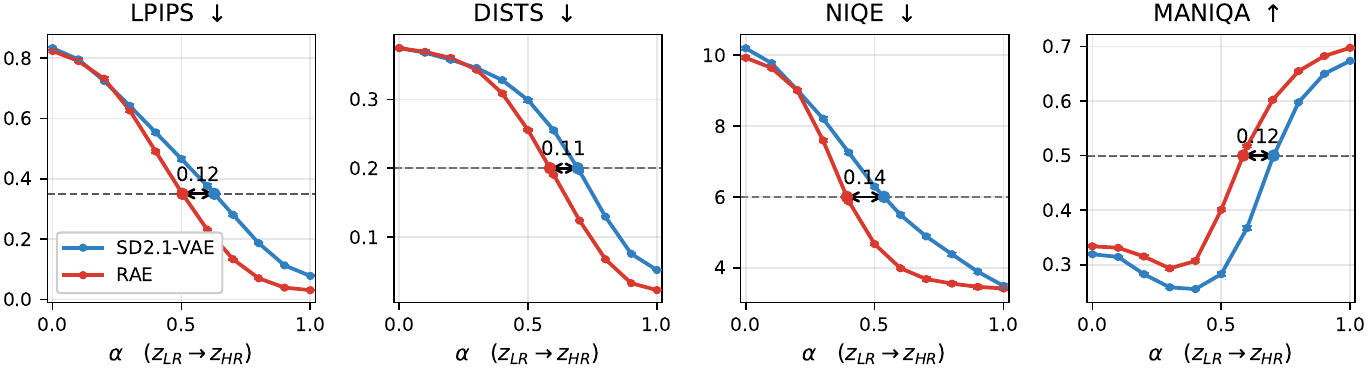}
\caption{\textbf{The latent line with the SD2.1 VAE}, on the same images and with the same protocol
and thresholds as Fig.~\ref{fig:interp}. The number above each arrow is the gap between the two
threshold crossings.}
\label{fig:interp-sd21}
\end{figure}

\paragraph{Endpoint matching and the SD2.1 replication.} SD3 is the primary comparison because its
endpoints match the RAE's. The two decode equally degraded images at $\alpha{=}0$ and near-identical
reconstructions at $\alpha{=}1$ (LPIPS $0.031$ vs $0.031$, DISTS $0.023$ vs $0.027$). We repeat the
study with the SD2.1 VAE on the same images (Fig.~\ref{fig:interp-sd21}). At $\alpha{=}0$ its decodes
are as degraded as the RAE's, and at $\alpha{=}1$ it matches the RAE on the perceptual metrics (NIQE
$3.49$ vs $3.42$, MANIQA $0.673$ vs $0.697$). Its reconstruction is less faithful to the pixels
(LPIPS $0.078$ vs $0.031$, PSNR $26.6$ vs $31.1$\,dB) because its latent has $4$ channels against the
$16$ of SD3, and fewer latent channels reconstruct less detail \citep{sd3}. This gap does not affect
the comparison. It lies at the end of the line, far below the thresholds (LPIPS $0.35$, DISTS $0.20$),
while the curves separate most in the middle of the line, where the crossings are decided. On all four
metrics the SD2.1 line crosses its threshold later than the RAE line, by $0.11$ to $0.14$ of the path,
and its LPIPS and NIQE crossings match the SD3 crossings almost exactly
(Table~\ref{tab:interp-thresholds}), so the conclusion does not rest on either VAE alone.

\paragraph{Random-direction control.} Walking from $z_{\mathrm{LR}}$ along a random direction of
the same per-image norm restores nothing in the fused latent. Over the whole walk NIQE moves from
$9.93$ to $9.95$ and LPIPS from $0.823$ to $0.801$, so early naturalness on the true line is a
property of the direction toward the clean embedding, not a sharpening bias of the decoder. On the fidelity side the control is strictly worse than the true path for
both latent spaces. The paired probability that the control's trajectory is worse is $1.00$ on LPIPS, DISTS, PSNR and SSIM. 

\paragraph{Noise augmentation is not the mechanism.} The RAE decoder is trained on noise-smoothed latents \citep{rae,raev2}, so its early naturalness could be suspected to come from this trained
robustness. Robustness is an invariance. Under a perturbation it keeps the decode what it already
was, which is exactly the behaviour of the random walk above; the true line instead changes the
decode toward the clean image, and an invariance cannot produce an improvement. The robustness is
also direction-blind, so it cannot explain why one specific direction, toward $z_{\mathrm{HR}}$,
improves every metric family at once (Fig.~\ref{fig:interp}).

\paragraph{Wrong-target control: membership is not fidelity.} A second control interpolates
$z_{\mathrm{LR}}$ toward the clean embedding of a \emph{different} image (a fixed derangement of
the manifest). Both decoders decode this wrong line to fully natural images (NIQE $3.3$--$3.4$ and
MUSIQ ${\approx}72$ at $\alpha{=}1$) while LPIPS to the true image never falls below $0.74$.
Reaching the manifold and reaching the right point on it are separate quantities, which is why
Fig.~\ref{fig:interp} reports one metric for each; it is also the geometric form of the
hallucination failure: a restorer can land on the manifold at the wrong point.

\section{Decoder channel ablation}
\label{app:channel}

This appendix backs \S\ref{sec:decoder-reads}. Every condition here is scored \emph{before} the
channel-mean alignment of \S\ref{sec:pipeline}, so the rows are internally comparable to one another
and offset from Table~\ref{tab:main}; the alignment is a per-channel shift and does not affect which
band the decoder depends on. The experiment is inference-only: the released
checkpoint is frozen throughout, and only the construction of the latent changes.

\paragraph{Protocol.}
In each ablated condition, eight of the latent's $23$ averaged layers are replaced, for every image, by
fixed \emph{training-set mean tokens}: per-layer, per-position means of shape $[23,1024,1024]$ computed
once over $512$ training images (seed $20260901$) under the evaluation preprocessing. Substituting the
mean rather than zero preserves the norm and coarse token statistics of the average while removing
image-specific content. In conditions B--D the broadcast term of layer $24$ is computed from the original tokens, so only the averaged band varies; condition E additionally replaces the broadcast term itself by the fixed vector $v = \overline{h_{24}}$, the training-set mean of the layer-$24$ spatial mean, obtained from the same mean-token tensor by linearity. Conditions: \textbf{B} no-shallow, layers $2$--$9$; \textbf{C} no-deep, layers $17$--$24$; \textbf{D} random eight-layer bands at three seeds ($\{3,10,11,14,15,17,18,24\}$, $\{4,5,6,10,16,17,19,20\}$, $\{3,4,7,8,10,11,13,23\}$); \textbf{E} wholesale deep removal, C's substitution plus the broadcast
replaced by $v$. All $3{,}443$ benchmark images are
re-scored per condition with the pipeline of \S\ref{sec:protocol}.

\begin{table}[t]
\centering
\renewcommand{\arraystretch}{0.95}
\footnotesize
\setlength{\tabcolsep}{4pt}
\caption{\textbf{Decoder channel ablation}, macro-averaged over the four benchmarks. B--D replace
eight of the $23$ averaged layers with training-set mean tokens at inference; E additionally replaces the broadcast term of layer $24$ by its training-set mean. D is
the mean of three random bands. \textbf{Bold} marks the cells the dissociation turns on: fidelity
collapsing without the shallow band and \emph{improving} without the deep one, further without the
broadcast.}
\label{tab:channel}
\resizebox{\textwidth}{!}{%
\begin{tabular}{lccccccccc}
\toprule
Condition & PSNR$\uparrow$ & SSIM$\uparrow$ & LPIPS$\downarrow$ & DISTS$\downarrow$ & NIQE$\downarrow$ & MUSIQ$\uparrow$ & MANIQA$\uparrow$ & TOPIQ$\uparrow$ & CLIP$\uparrow$ \\
\midrule
A unmodified & 23.72 & 0.6265 & 0.3012 & 0.1971 & 4.320 & 70.45 & 0.6717 & 0.6820 & 0.7059 \\
B no-shallow (layers $2$--$9$) & \textbf{16.43} & 0.5179 & 0.5194 & 0.3807 & 5.021 & 70.66 & 0.6585 & 0.6454 & 0.5994 \\
C no-deep (layers $17$--$24$) & \textbf{24.60} & 0.6565 & 0.3079 & 0.2088 & 4.687 & 66.75 & 0.6134 & 0.6177 & 0.6622 \\
D random band (3-seed mean) & 20.48 & 0.6069 & 0.3728 & 0.2733 & 4.884 & 69.59 & 0.6402 & 0.6497 & 0.6482 \\
\quad seed 0 & 22.67 & 0.6356 & 0.3270 & 0.2308 & 4.814 & 68.25 & 0.6230 & 0.6432 & 0.6541 \\
\quad seed 1 & 19.90 & 0.6030 & 0.3777 & 0.2820 & 4.931 & 70.19 & 0.6500 & 0.6531 & 0.6572 \\
\quad seed 2 & 18.87 & 0.5821 & 0.4137 & 0.3070 & 4.907 & 70.34 & 0.6476 & 0.6527 & 0.6332 \\
E no-deep $+$ no-broadcast & \textbf{24.99} & 0.6642 & 0.3237 & 0.2220 & 4.746 & 63.35 & 0.5606 & 0.5639 & 0.6205 \\

\bottomrule
\end{tabular}}
\end{table}

\paragraph{Results.}
Table~\ref{tab:channel} gives the macro-averages. Define, per condition, $F$ as the mean relative
degradation over PSNR/SSIM/LPIPS and $P$ over NIQE/MUSIQ/MANIQA/CLIP-IQA, both signed so that positive
is worse. B gives $F = +40.2\%$ against $P = +8.2\%$; C gives $F = -2.1\%$, a net
\emph{improvement}, against $P = +7.2\%$; the three random bands give $F - P$ of $-3$, $+9$ and
$+15$ points. B's separation ($+31.9$) and C's reversed separation ($-9.2$) both lie outside the
random range, and the crossing holds on each of the four benchmarks separately. $P$ is computed over the perceptual metrics plotted in Fig.~\ref{fig:intervention}; the table also reports DISTS and TOPIQ. Fig.~\ref{fig:intervention} (right) shows what
the two failure modes look like on a deterministically chosen sample.

\paragraph{The broadcast term.}
Removing the broadcast term together with the deep band, condition E, gives $F = -1.3\%$ (PSNR
$+0.76$--$1.79$\,dB on every benchmark, SSIM $+0.038$) against $P = +12.1\%$, a separation of $13.4$
points, wider than C's $9.2$, with the crossing again holding per benchmark.

\section{The adversarial term: schedule}
\label{app:adv}
\S\ref{sec:critic} describes the critic; this appendix gives its schedule. Stage 2 runs the critic
on the full $512$px image with one update per generator step. Three scheduling choices in the final
stage matter more than the loss weights. \emph{(i)} The critic sees a $496{\times}496$ centre crop
while every other loss sees the full $512$px image, so that DINOv3-B's $16$px patch grid is offset by half a patch from the decoder's $16$px token grid. The decoder produces each $16$px patch from one token, which can leave faint seams on the patch borders. If the two grids coincided, every seam would lie on a critic patch border, at the same position in every patch, and the critic would learn it as part of a real image. With the offset, each seam runs through the middle of a critic patch, where the critic detects it and pushes the decoder to remove it. \emph{(ii)} The critic runs on a two-time-scale
update rule \citep{ttur} (a $6\times$ higher learning rate and two updates per generator step), since
a frozen-feature critic that lags the generator supplies no usable gradient. \emph{(iii)} The
adversarial term is gated: the critic is warmed at $\lambda_{\mathrm{adv}}\!=\!0$ and the weight then
ramped in linearly, so the decoder settles on reconstruction before the critic engages.

\section{RAESR-Distilled}
\label{app:distill}

Because the encoder and the trained decoder are both frozen at test
time, the encoder ($303$M, $42\%$ of the model) can be replaced without retraining anything. We
distil it into a frozen DINOv3-S/16$+$ ($28.7$M) plus a learned $57.8$M projector, trained to
predict the latent of Eq.~\ref{eq:agg}, so the released decoder is reused unchanged. Its training protocol is in App.~\ref{app:training}.

\section{The substrate ablation of \S\ref{sec:substrate}}
\label{app:substrate}

This appendix gives the protocol and the full scores of the substrate ablation in \S\ref{sec:substrate}, in which the frozen DINOv3-L is replaced by the frozen SD3 VAE and everything else is kept.

\paragraph{Protocol.}
Setup: the SD3
\texttt{AutoencoderKL} runs frozen in deterministic posterior mode with its config scaling; the
$[16,64,64]$ latent is rearranged by parameter-free pixel-unshuffle onto the shared $32{\times}32$
grid as $1{,}024$ tokens of width $64$ (all $65{,}536$ scalars preserved), and only the decoder's
input projection changes shape ($1152{\times}64$, re-initialised; every other weight warm-starts from
the official RAEv2 ViT-XL decoder, the initialisation the released model also uses). Training repeats the three-stage chain: $30$k reconstruction steps (L1 $+$
LPIPS, multi-resolution), $50$k adversarial steps on LSDIR-$512$ under VOSR-strong degradation with
the frozen DINOv3-B critic; and then stage 3 exactly as released: seed $42$, $600$ steps, global
batch $64$, mixed $30/50/20$ severity, ramped adversarial weight, and \emph{zero} weight on every
no-reference quality loss, matching the released model's recipe term for term. Both arms are the final $600$-step EMA with no selection, and both are scored before the channel-mean alignment of
\S\ref{sec:pipeline}, so the pair is internally comparable and offset from Table~\ref{tab:main}.

\paragraph{Scores.} Table~\ref{tab:substrate} gives the full scores. The per-benchmark rows locate the failure: the reference-perceptual gap is largest on LSDIR (LPIPS
$+0.081$, DISTS $+0.056$), the most heavily degraded set, and the sweep is total. The deviations from the released recipe are minor. Stage 1 was trained at multiple resolutions on the full $166{,}994$-image training pool rather than on the $512$px LSDIR set, and stage 2 ran at global batch $15$ rather than $16$. 

\begin{table}[!ht]
\centering
\renewcommand{\arraystretch}{0.92}
\footnotesize
\setlength{\tabcolsep}{3.4pt}
\caption{\textbf{Encoder-substrate ablation} under the released recipe; the DINOv3 substrate
leads on all nine metrics.}
\label{tab:substrate}
\resizebox{\textwidth}{!}{%
\begin{tabular}{lccccccccc}
\toprule
 & PSNR$\uparrow$ & SSIM$\uparrow$ & LPIPS$\downarrow$ & DISTS$\downarrow$ & NIQE$\downarrow$ & MUSIQ$\uparrow$ & MANIQA$\uparrow$ & TOPIQ$\uparrow$ & CLIP$\uparrow$ \\
\midrule
\multicolumn{10}{l}{\emph{Released recipe (no quality metric in the loss)}} \\
RAESR (released) & \textbf{23.72} & \textbf{0.6265} & \textbf{0.3012} & \textbf{0.1971} & \textbf{4.320} & \textbf{70.45} & \textbf{0.6717} & \textbf{0.6820} & \textbf{0.7059} \\
SD3-VAE substrate & 23.61 & 0.6033 & 0.3734 & 0.2495 & 4.344 & 69.76 & 0.6300 & 0.6632 & 0.7011 \\
\quad on LSDIR & 19.73 & 0.4242 & 0.4291 & 0.2481 & 3.274 & 73.33 & 0.6492 & 0.7023 & 0.7210 \\
\quad on RealSR & 24.25 & 0.6831 & 0.3390 & 0.2521 & 4.694 & 69.26 & 0.6313 & 0.6481 & 0.6801 \\
\quad on DRealSR & 27.15 & 0.7320 & 0.3711 & 0.2618 & 5.528 & 65.82 & 0.6052 & 0.6286 & 0.6776 \\
\quad on DIV2K-Val & 23.32 & 0.5737 & 0.3546 & 0.2362 & 3.878 & 70.62 & 0.6345 & 0.6738 & 0.7259 \\
\bottomrule
\end{tabular}}
\end{table}

\begin{figure}[!ht]
\centering
\includegraphics[width=\textwidth]{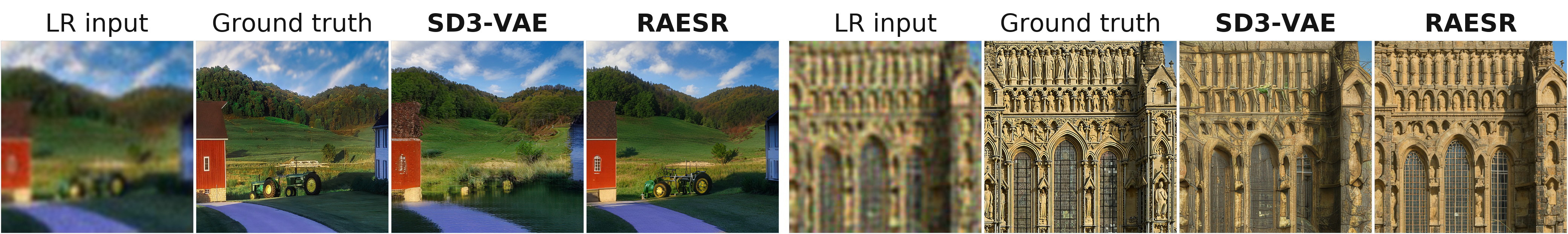}
\caption{\textbf{Swapping the substrate: two further examples.} Format as in
Fig.~\ref{fig:substrate}. The tractor becomes a pond and the cathedral's tracery is redrawn as
generic stonework.}
\label{fig:substrate-extra}
\end{figure}

\paragraph{Examples.} Fig.~\ref{fig:substrate-extra} shows two further examples.

\section{Training details and protocols}
\label{app:training}

\paragraph{Training data.}
Stages 1 and 2 use the LSDIR training split ($84{,}991$ images). Stage 3
uses a row-weighted manifest of LSDIR ($\times10$), FFHQ-1024 ($\times8$) and Aesthetic-4K
($\times1$): $1{,}421{,}913$ rows over $166{,}994$ unique images. In stages 2 and 3 every image is
Lanczos-resized to shortest side $512$ and randomly cropped to $512^2$. The evaluation benchmarks are
disjoint by construction: the LSDIR test set is the official $250$-image validation split, and
DIV2K, RealSR and DRealSR contribute no images to any stage. Degradation is synthesised on the fly
with no fixed LR set. In stages 2 and 3 the clean crop is USM-sharpened before the degradation
chain, following Real-ESRGAN's data synthesis, while the unsharpened crop remains the target of
every loss and the critic's real image; the $128^2$ result is quantised to $8$ bits and bicubically
upsampled to $512$ before the encoder.

\paragraph{Common settings.}
Only the decoder trains; the DINOv3-L/16 encoder is frozen in every stage. AdamW with
$\beta=(0.9,0.95)$, no weight decay, no gradient clipping, bf16 autocast, seed $42$, decoder EMA
$0.9978$ updated every step; selected checkpoints and all evaluation use the EMA weights. Losses are
$\mathcal{L}_1$, VGG16 LPIPS at weight $1.0$, and where active the relativistic adversarial term of
\S\ref{sec:critic} with finite-difference $R_1/R_2$ ($\lambda_{\mathrm{gp}}=0.25$, $\epsilon=0.01$)
and logit centring ($\lambda_{\mathrm{cp}}=0.01$); no instance noise and no other loss terms.

\paragraph{Stage 1: bicubic initialisation.}
Initialised from the official RAEv2 ViT-XL reconstruction decoder \citep{raev2}, its original
positional embedding replaced by 2D RoPE, as is every ablation in this paper.
Each step resizes to shortest side $512$, crops $512^2$, and forms the input by bicubic
$\times4$ down- and up-sampling; no other degradation. Reconstruction losses only.
Learning rate $2\times10^{-5}$, $1{,}000$-step warmup, cosine to $2\times10^{-6}$ over a
$50{,}000$-step horizon; the EMA at update $30{,}000$ is carried forward.

\paragraph{Stage 2: strong degradation at $512$.}
From the stage-1 EMA, four GPUs $\times$ batch $4$ (global $16$), $50{,}000$ steps, sampled with
replacement. Degradation is the VOSR-strong two-order profile. Reconstruction losses only for
$2{,}000$ steps, then $\lambda_{\mathrm{adv}}=0.5$ as a step switch. Critic: frozen DINOv3-B/16, one
convolutional head per block ($12$ heads, $191$M trainable parameters in total, discarded after training),
fresh initialisation, full $512^2$ input, one update per generator step. Learning rates
$2\times10^{-5}$ (decoder) and $1\times10^{-4}$ (critic), $1{,}000$-step warmup, cosine to
$2\times10^{-6}$ over $50{,}000$. The final EMA is carried forward.

\paragraph{Stage 3: mixed-severity finish (released).}
From the stage-2 EMA with freshly initialised critic heads, two GPUs $\times$ batch $32$ (global
$64$), $600$ steps sampled without replacement ($38{,}400$ crops). Each image draws its severity:
VOSR-strong $0.30$, StableSR-medium $0.50$, mild single-order $0.20$. Reconstruction losses only
for $200$ steps; the adversarial term then ramps linearly toward $1.0$ over $800$ steps, so it stands
at $\approx0.5$ when the run ends at step $600$. Critic as in stage 2 but on a $496^2$ centre crop and
two updates per generator step. Learning rates $1\times10^{-4}$ (decoder) and $6\times10^{-4}$
(critic), $200$-tick warmup and cosine to $2\times10^{-6}$ over a $2{,}778$-tick horizon, the
scheduler ticking twice per step. The released model is the EMA at $600$, with no retrospective selection.

\paragraph{Distillation.}
Two stages, seed $20260812$, encoder and decoder frozen throughout; only the $57.8$M projector (a
ViT of width $768$, depth $8$, $12$ heads with 2D RoPE) trains. Stage 1 fits the projector to the
teacher aggregate under latent MSE alone (learning rate $10^{-5}$; the step-$19{,}500$ checkpoint is carried forward). Stage 2
runs $20{,}000$ steps at global batch $65$ (five ranks $\times$ $13$), learning rate
$3\times10^{-6}$, $100$-step warmup, gradient clipping at $1.0$, minimising
$\mathrm{MSE}_{\mathrm{latent}} + 0.5\,\mathcal{L}_1 + 1.0\,\mathcal{L}_{\mathrm{LPIPS}}$ where the
image terms compare the student's and the released decoder's outputs. The released student is the
final step-$20{,}000$ checkpoint, which was also the best on the held-out validation objective.

\section{Sample gallery}
\label{app:gallery}
Five selected samples per benchmark, each shown with all sixteen panels at their original resolution.

\begin{figure}[p]
\centering
\includegraphics[width=0.78\textwidth]{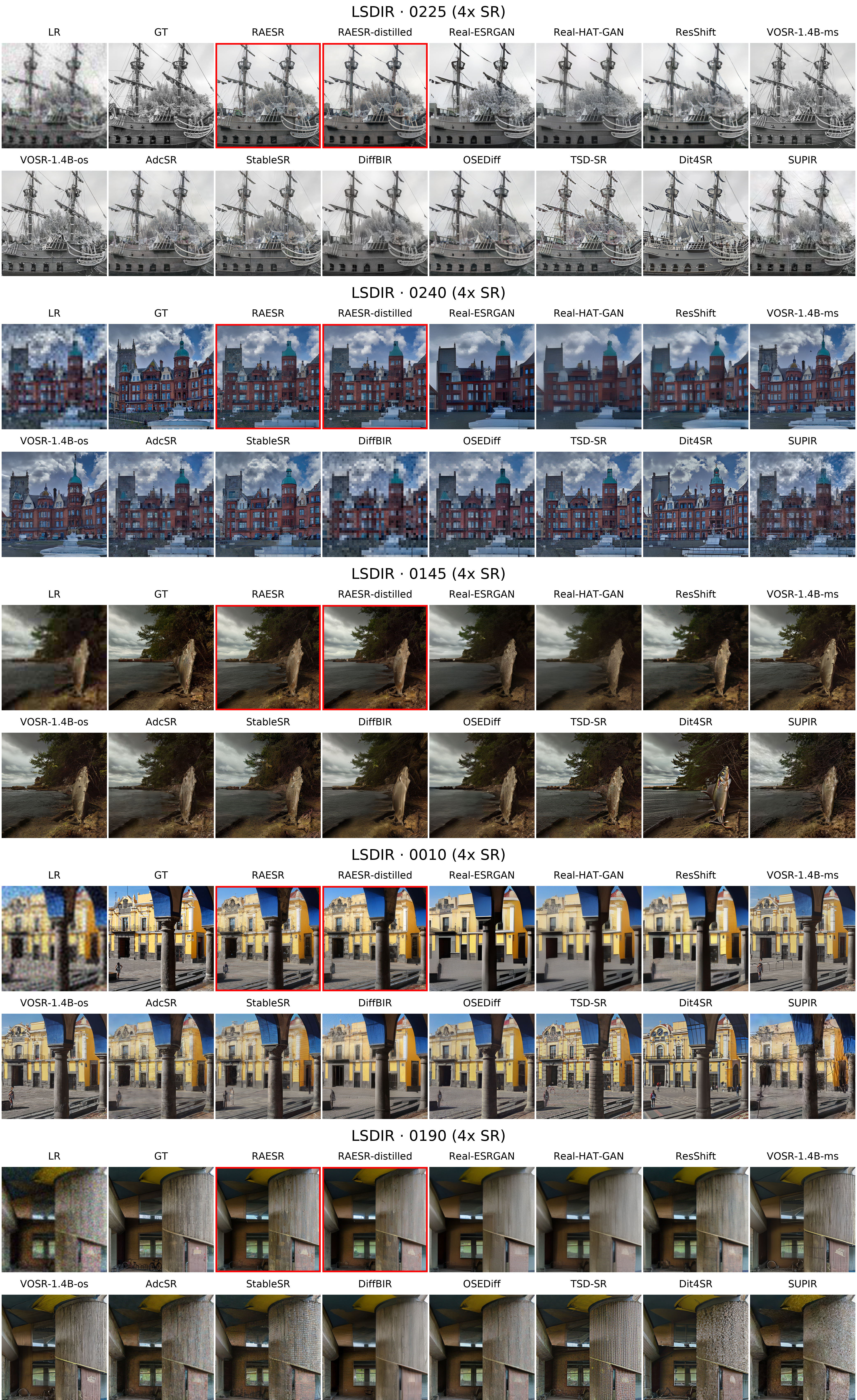}
\caption{\textbf{LSDIR.} The first sample is the example of Fig.~\ref{fig:qualitative}. \textbf{Best viewed zoomed in.}}
\label{fig:gal-1}
\end{figure}
\clearpage

\begin{figure}[p]
\centering
\includegraphics[width=0.78\textwidth]{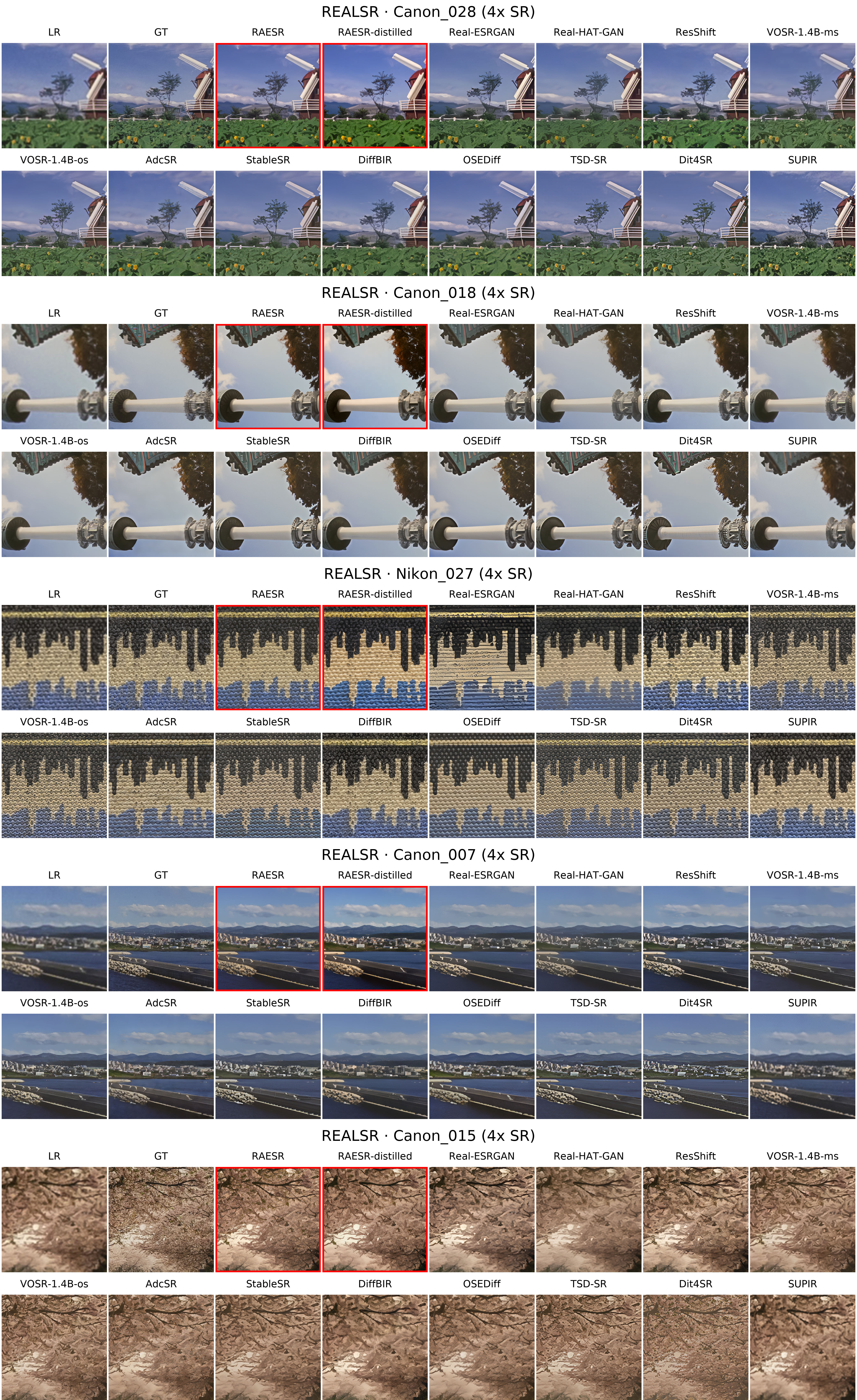}
\caption{\textbf{RealSR.} The first sample is the example of Fig.~\ref{fig:qualitative}. \textbf{Best viewed zoomed in.}}
\label{fig:gal-2}
\end{figure}
\clearpage

\begin{figure}[p]
\centering
\includegraphics[width=0.78\textwidth]{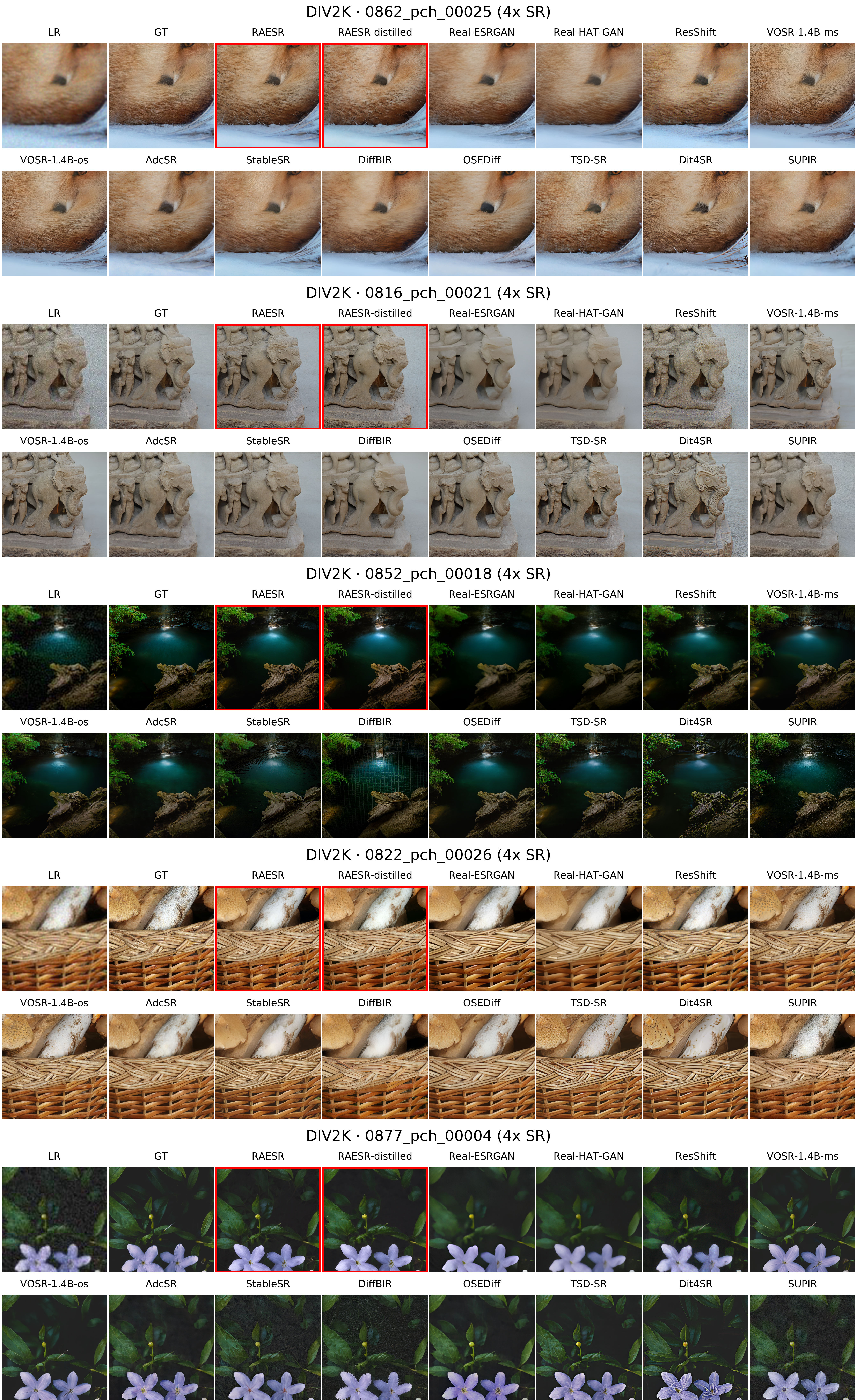}
\caption{\textbf{DIV2K-Val.} The first sample is the example of Fig.~\ref{fig:qualitative}. \textbf{Best viewed zoomed in.}}
\label{fig:gal-3}
\end{figure}
\clearpage

\begin{figure}[p]
\centering
\includegraphics[width=0.78\textwidth]{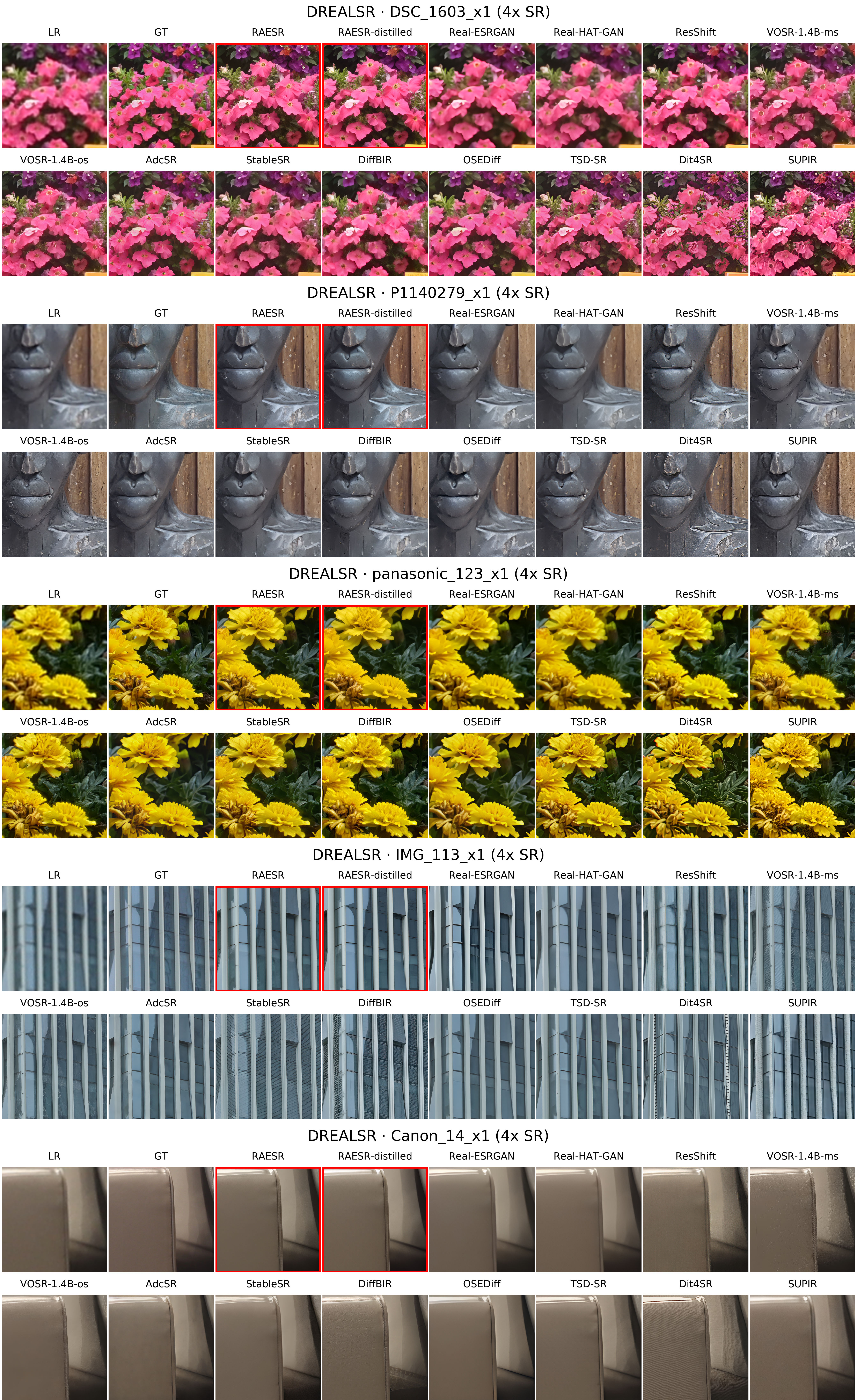}
\caption{\textbf{DRealSR.} \textbf{Best viewed zoomed in.}}
\label{fig:gal-4}
\end{figure}
\clearpage

\end{document}